\documentclass{article}

\newif\ifneurips
\neuripsfalse      %

\PassOptionsToPackage{numbers,sort&compress}{natbib}

\ifneurips
    \usepackage{neurips_2025}

\else
    \usepackage[preprint]{neurips_2025}
\fi

\usepackage[utf8]{inputenc} %
\usepackage[T1]{fontenc}    %
\usepackage{hyperref}       %
\usepackage{url}            %
\usepackage{booktabs}       %
\usepackage{amsfonts}       %
\usepackage{nicefrac}       %
\usepackage{microtype}      %
\usepackage{xcolor}         %

\usepackage{amsfonts}
\usepackage{amsmath}
\usepackage{xfrac}
\usepackage{bm}
\usepackage{bbold}
\usepackage{mathtools}
\usepackage{wasysym}
\usepackage{amsthm}

\theoremstyle{plain}

\usepackage{algorithm}
\usepackage{algpseudocode}
\usepackage{algorithmicx}

\usepackage{tcolorbox}

\usepackage{booktabs}
\usepackage{array}
\usepackage{longtable}
\DeclareMathOperator*{\argmin}{arg\,min}					%
\newcommand{\E}{\mathbb{E}} 		                        %

\DeclareMathAlphabet{\mathsfit}{\encodingdefault}{\sfdefault}{m}{sl}
\SetMathAlphabet{\mathsfit}{bold}{\encodingdefault}{\sfdefault}{bx}{n}

\newcommand{\inp}{\mathbf{x}}
\newcommand{\target}{\mathbf{t}}

\newcommand{\out}{\mathbf{y}}

\newcommand{\param}{\boldsymbol{\theta}}

\newcommand{\oparam}{\param^{\star}}

\newcommand{\pp}[2]{\frac{\partial #1}{\partial #2}}
\newcommand{\dd}[2]{\frac{\mathrm{d}#1}{\mathrm{d}#2}}

\newcommand{\traingradwithinputs}[2]{\nabla_{\params}\mathcal{L} (#1, #2)}

\newcommand{\traingrad}{\nabla_{\params}\mathcal{L} (\oparam, \dataPoint_m)}
\newcommand{\traingradwithfinalparams}{\nabla_{\params}\mathcal{L} (\finalParams(\xi), \dataPoint_m)}
\newcommand{\traingradwithfinalparamsnorand}{\nabla_{\params}\mathcal{L}(\finalParams, \dataPoint_m)}
\newcommand{\querygrad}{\nabla_{\params}f_{\dataPoint_q}(\oparam)}
\newcommand{\querygradwithfinalparams}{\nabla_{\params}f_{\dataPoint_q} (\finalParams(\xi))}
\newcommand{\querygradwithfinalparamsnorand}{\nabla_{\params}f_{\dataPoint_q} (\finalParams)}

\newcommand{\ggn}{\mathbf{G}}

\newcommand{\invhessian}{\hessian^{-1}}
\newcommand{\testdata}{\mathcal{D}_{\text{query}}}
\newcommand{\preconditioner}{\mathbf{P}}
\newcommand{\fisher}{\mathbf{F}}

\newcommand{\algoname}{ASTRA}
\newcommand{\algonameif}{\algoname-IF}
\newcommand{\source}{SOURCE}
\newcommand{\algonamesource}{\algoname-\source}
\newcommand{\ekfac}{EKFAC}

\newcommand{\ekfacif}{\ekfac-IF}
\newcommand{\ekfacsource}{\ekfac-\source}

\newcommand{\ihvp}{iHVP}
\newcommand{\ihvps}{iHVPs}

\newcommand{\lissa}{LiSSA}

\newcommand{\conjugategradient}{CG}

\newcommand{\cifarten}{CIFAR-10}
\newcommand{\resnetnine}{ResNet-9}
\newcommand{\gpttwo}{GPT-2}
\newcommand{\wikitexttwo}{WikiText-2}
\newcommand{\mnist}{MNIST}
\newcommand{\fashionmnist}{FashionMNIST}
\newcommand{\trak}{TRAK}
\newcommand{\logra}{LOGRA}
\newcommand{\vni}{NI}
\newcommand{\sni}{SNI}

\newcommand{\trainingData}{\mathcal{D}}

\newcommand{\dataPoint}{\boldsymbol{z}}
\newcommand{\params}{\boldsymbol{\theta}}
\newcommand{\optParams}{\boldsymbol{\theta}^\star}

\newcommand{\finalParams}{\boldsymbol{\theta}^s}

\newcommand{\loss}{\mathcal{L}}
\newcommand{\cost}{\mathcal{J}}

\newcommand{\eye}{\mathbf{I}}
\newcommand{\LR}{\eta}

\newcommand{\hessian}{\mathbf{H}}

\newcommand{\segment}{\mathbf{S}}

\newif\ifshowappendix    %
\showappendixfalse        %

\usepackage{hyperref}
\definecolor{customblue}{rgb}{0,0.08,0.45}
\definecolor{customgreen}{RGB}{85,107,47}
\definecolor{custompink}{RGB}{170,51,106}
\definecolor{egyptianblue}{rgb}{0.06, 0.2, 0.65}
\usepackage[nameinlink]{cleveref}
\creflabelformat{equation}{#2#1#3}
\hypersetup{
    colorlinks=true,
    linkcolor=customblue,
    filecolor=magenta,
    urlcolor=custompink,
    citecolor=egyptianblue
} 

\usepackage{caption}
\usepackage{wrapfig}
\usepackage{soul}
\setul{0.5ex}{0.3ex}
\newcommand{\ulcolor}[2][Red]{\setulcolor{#1}\ul{#2}}
\usepackage{pifont}
\newcommand{\cmark}{\ding{51}} 
\newcommand{\xmark}{\ding{55}}

\title{Better Training Data Attribution via \\ Better Inverse Hessian-Vector Products}

\author{
  Andrew Wang\thanks{Correspondence to andrewwang@cs.toronto.edu.}\;\textsuperscript{\normalfont 1,2} \quad Elisa Nguyen\textsuperscript{\normalfont 3} \quad Runshi Yang\textsuperscript{\normalfont 1,2}\\ \textbf{Juhan Bae\textsuperscript{\normalfont 1,2} \quad  Sheila A. McIlraith\textsuperscript{\normalfont 1,2,4} \quad Roger Grosse\textsuperscript{\normalfont 1,2,4}} \\
  \textsuperscript{\normalfont 1}University of Toronto \quad
  \textsuperscript{\normalfont 2}Vector Institute for Artificial Intelligence \\ \textsuperscript{\normalfont 3}T\"ubingen AI Center, University of T\"ubingen \\ \textsuperscript{\normalfont 4}Schwartz Reisman Institute for Technology and Society
}

\begin{document}

\maketitle
\begin{abstract}
Training data attribution (TDA) provides insights into which training data is responsible for a learned model behavior. Gradient-based TDA methods such as influence functions and unrolled differentiation both involve a computation that resembles an inverse Hessian-vector product (\ihvp), which is difficult to approximate efficiently. 
We introduce an algorithm (\algoname) which uses the \ekfac-preconditioner on Neumann series iterations to arrive at an accurate \ihvp \space approximation for TDA. \algoname \space is easy to tune, requires fewer iterations than Neumann series iterations, and is more accurate than \ekfac-based approximations. Using \algoname, we show that improving the accuracy of the \ihvp \space approximation can significantly improve TDA performance.

\end{abstract}

\setcounter{footnote}{0}
\section{Introduction}
Machine learning systems derive their behavior from the data they are trained on. \textit{Training data attribution} (TDA) is a family of techniques that help uncover how individual training examples influence model predictions. As such, TDA is a valuable tool with applications in data valuation and curation~\citep{choe2024your, jain2024improving, teso2021interactive, hara2019data}, interpreting model behavior~\citep{koh2019accuracy, yeh2018representer, grosse2023studyinglargelanguagemodel, akyurek2022tracing, wang2023error, ilyas2022datamodels, mlodozeniec2024influence}, building more equitable and transparent machine learning systems~\citep{brunet2019understanding, wang2024fairif} and investigating questions of intellectual property and copyright by tracing outputs back to specific data sources~\citep{van2021memorization, mlodozeniec2024influence, mezzi2025owns}, among other applications.

Influence functions (IF) \citep{hampel1974influence, koh2017understanding} and unrolled differentiation \citep{hara2019data, chen2021hydra, bae2024training, wang2024capturing, ilyas2025magic} are two gradient-based TDA methods that involve, or can be approximated as computing inverse Hessian-vector products (\ihvps).\footnote{Other gradient-based TDA methods such as  \trak \space \citep{park2023trak} or \logra \space\citep{choe2024your} also involve \ihvps \space but use additional techniques such as random/PCA gradient projection.} Inverting the Hessian is infeasible but for the smallest of neural networks, so the \ihvp \space is typically computed without explicitly constructing the Hessian. There are a number of choices available: \citet{koh2017understanding}, who first introduced influence functions to deep learning, use the iterative algorithm \lissa \space \citep{agarwal2017second}, which is a method based on stochastic Neumann series iterations\footnote{\lissa \space was introduced in the context of optimization and contains other components, but we refer to the component that computes the \ihvp, as found in the \citet{koh2017understanding} implementation. Henceforth, we will use the terms \lissa \space and \sni \space interchangeably, with their slight difference described in \Cref{appendix:extended_preliminaries}.} (\sni)
 \citep{horn2012matrix, lorraine2020optimizing} that can take \textit{thousands} of iterations to converge to an unbiased solution \citep{koh2017understanding, grosse2023studyinglargelanguagemodel}. Alternatively, \citet{grosse2023studyinglargelanguagemodel} adopt a tractable parametric approximation to the Hessian \citep{bae2024training, choe2024your, mlodozeniec2024influence} with \ekfac \space \citep{martens2015optimizing, george2018fast}. \ekfac \space makes several simplifying assumptions that hold only approximately in practice, but they dramatically lower both computational and memory cost, making it feasible to scale to billion‑parameter language models \citep{grosse2023studyinglargelanguagemodel}. However, \ekfac \space influence functions (\ekfacif) computed on converged models only correlate modestly with ground truth over a variety of datasets and model architectures \citep{bae2024training} and tend to struggle for architectures involving convolution, suggesting further room for improvement. We aim to improve \ekfac-based TDA methods in this paper by improving the \ihvp \space approximation.

\definecolor{colpresto}{RGB}{0,114,178}
\definecolor{colekfac}{RGB}{0,158,115}
\definecolor{colsni}{RGB}{213,94,0}
\begin{wrapfigure}{l}{0.50\textwidth}
    \centering
    \includegraphics[width=\linewidth]{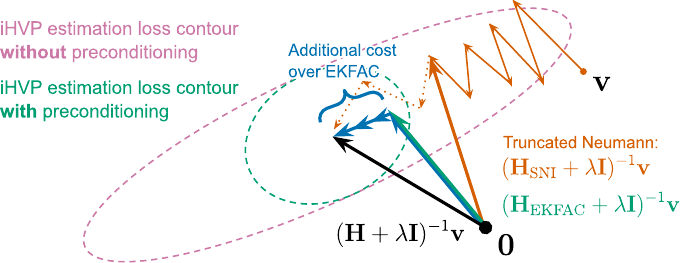}
    \captionsetup{font=small}
    \caption{\small{The objective is to compute the damped \protect \ihvp \space $(\mathbf{H}+ \lambda \mathbf I)^{-1}\mathbf v$. Preconditioning Stochastic Neumann Iterations (SNI) with \protect \ekfac \space (\ulcolor[colpresto]{ASTRA}) improves the convergence speed of the \protect \ihvp \space approximation. Initialized at $\mathbf{0}$, it results in the same approximation as using \protect \ulcolor[colekfac]{EKFAC} after one iteration. 
    \ulcolor[colsni]{SNI} may require thousands of iterations to converge, and truncating early results in undesirable implicit damping.}}
    \label{fig:fig1}
\end{wrapfigure}

Computing \ihvps \space in the context of TDA can be seen as finding the minimizer to high-dimensional quadratic optimization problems in parameter space \citep{nocedal1999numerical, vicol2022implicit,lorraine2020optimizing}. Curvature matrices for large neural networks are known to be ill-conditioned \citep{sagun2016eigenvalues, ghorbani2019investigation}, causing slow convergence for iterative methods such as \sni. While costly and rather difficult to tune \citep{koh2017understanding, klochkov2024revisiting, schioppa2022scaling}, the upside of \sni \space is that it produces a \textit{consistent} estimator -- the algorithm converges to the \ihvp \space in the limit as more compute is used \citep{agarwal2017second}. In contrast, while \ekfac \space often provides a better cost vs. accuracy tradeoff for TDA \citep{grosse2023studyinglargelanguagemodel}, there is no simple way to improve its accuracy by applying more compute. Our central insight is that we can combine the best of both worlds: we can repurpose the \ekfac \space decomposition --- which needs to be computed for \ekfac‑based influence functions and unrolled differentiation anyways --- as a preconditioner for \sni, yielding a cost effective procedure for improving the iHVP accuracy (\Cref{fig:fig1}).  Our contributions are as follows: 

First, we present an algorithm called \algoname \space which uses \ekfac \space as a preconditioner for \sni \space with the aim of computing cost-effective and accurate \ihvps. \algoname \space can be applied to both influence functions (\algonameif) and unrolled differentiation via an approximation called \source \space (\algonamesource) \citep{bae2024training} among other applications. To the best of our knowledge, no prior work has applied the \ekfac \space preconditioner to \sni \space for the TDA setting. In our experiments, the incremental cost of \algonameif \space and \algonamesource \space  was only \textit{hundreds} of iterations, compared to \ekfacif \space  and \ekfacsource, respectively. In contrast to other settings in which we would encounter \ihvps, we are often willing to pay this extra computational cost to obtain a more accurate \ihvp \space for TDA.

While past papers have questioned the reliability of influence functions \citep{basu2020influence, schioppa2024theoretical}, we show that influence functions computed with accurate \ihvps \space have strong predictive power, even in settings in which the assumptions in its derivations may not hold \citep{bae2022if, schioppa2024theoretical}. In our experiments, \algonameif \space was able to achieve a Spearman correlation score \citep{spearman1904proof} with ground-truth retraining of around 0.5 across many settings,\footnote{More precisely, we use a widely-used evaluation metric called Linear Datamodeling Score (LDS) \citep{park2023trak}, which measures the rank correlation between ground-truth retraining outcomes and a TDA algorithm's predictions. LDS is defined and performance comparisons are provided in \Cref{section:experiments}.} and ensembling these predictions often raised performance to 0.6. \algoname \space provides an accurate approximation of the \ihvp, which significantly increases the efficacy of ensembling \citep{park2023trak, bae2024training} compared to \ekfac \space and also significantly improves TDA performance for architectures involving convolution layers. For the experiment involving a convolution architecture, performance of ensembled \algonameif \space was 0.6, a large increase from \ekfacif's 0.25.

Finally, truncating Neumann series has an implicit damping effect \citep{vicol2022implicit, gerfo2008spectral}, which has a disproportional impact on low curvature directions. To quantify the downstream impact on influence functions performance, we perform an ablation study by using the \ekfac \space eigendecomposition to project the \ihvp \space onto subspaces containing different levels of curvature during Neumann series iterations. We show that these low curvature components are essential for high quality influence estimates. This suggests that good influence function performance demands careful hyperparameter tuning, which can be costly, especially for off-the-shelf iterative solvers. In contrast, \algoname \space requires much less hyperparameter tuning; we use a set of hyperparameters determined by simple heuristics which worked well for all of our experiments. We also note that while \ekfac \space was introduced as a preconditioner in second‑order optimization \citep{martens2015optimizing}, its compact representation of the eigendecomposition remains relatively underappreciated --- the method in this ablation study in which we use the \ekfac \space eigendecomposition to analyze the quadratic objective could therefore be of independent interest.

\section{Preliminaries}\label{background}
Given a training dataset $\mathcal{D} = \{ \dataPoint_i \}_{i=1}^N$ where $\dataPoint_i = (\inp^{(i)}, \target^{(i)})$ is an input-target pair, and a model parameterized by $\params \in \mathbb{R}^D$, let $g( \params, \inp^{(i)})$ denote the model output on $\inp^{(i)}$, let $\mathcal{L}(\out,\target)$ be a convex loss function. We define a training objective $\cost(\params, \mathcal{D}) \coloneqq \frac{1}{N} \sum_{i =1} ^N \mathcal{L}(g( \params, \inp^{(i)}),\target^{(i)})$ as the average loss over $\mathcal{D}$. Given a query data point $\dataPoint_q$ and a measurement function $f_{\dataPoint_q}(\params)$, such as correct-class margin \citep{park2023trak}, an \textit{idealized} objective of TDA is to approximate the impact of removing a training example $\dataPoint_m$ from the training dataset $\mathcal{D}$ on $f_{\dataPoint_q}$. A pointwise TDA method $\tau$  assigns a score $\tau(\dataPoint_m, \dataPoint_q, \mathcal{D})$ that measures the impact of removing $\dataPoint_m$ from $\mathcal{D}$ on $f_{\dataPoint_q}$. Since modern neural networks often exhibit multiple optima, the stochasticity in the optimization process from sources such as parameter initialization \citep{glorot2010understanding}, sampled
dropout masks \citep{srivastava2014dropout}, and mini-batch ordering \citep{li2014efficient} can result in different learned optima, which we denote $\finalParams$. Let $\xi$ be a random variable which captures this stochasticity in the training procedure \citep{epifano2023revisiting, nguyen2024bayesian}. For a full list of notation and acronyms, see \Cref{appendix:notation}. We discuss preliminaries briefly, with further details in \Cref{appendix:extended_preliminaries}.
 
\subsection{Computing Inverse Hessian-Vector Products for TDA}\label{subsection:compute_ihvps}
Inverse Hessian-vector products (\ihvps) are ubiquitous in many machine learning settings beyond TDA \citep{lorraine2020optimizing, martens2015optimizing, maclaurin2015gradient, ritter2018scalable}, such as second-order optimization \citep{tieleman2012lecture, martens2015optimizing, kingma2014adam}, but different settings have different considerations when trading off cost vs. accuracy. The canonical second-order optimization method -- Newton's method -- utilizes an inverse Hessian-gradient product to compute the Newton step, and may achieve faster local convergence than first-order methods \citep{nocedal1999numerical}. %
However, computing an \ihvp \space is substantially more expensive than a standard gradient computation. In optimization, there exists a tradeoff between devoting extra compute to obtain a better \ihvp \space approximation versus taking more steps in the optimization procedure \citep{byrd2016stochastic}. Because most deep learning optimizers rely on stochastic updates, the extra cost of a highly precise curvature estimate often is not justified and popular optimizers such as Adam \citep{kingma2014adam} default to the much cheaper diagonal preconditioner. In contrast, we are typically willing to pay a higher cost for accurate \ihvp \space approximations in the TDA context, since -- as we will show --  an accurate \ihvp \space may significantly improve TDA performance.

Two popular ways of computing the \ihvp \space for TDA are \ekfac \space \citep{martens2015optimizing, george2018fast, grosse2023studyinglargelanguagemodel} and \lissa \space \citep{agarwal2017second, koh2017understanding}. %
\ekfac \space makes a block-diagonal parametric approximation of the Fisher Information Matrix\footnote{For standard regression and classification tasks whose outputs can be seen as the natural parameters of an exponential family, the Fisher Information Matrix $\fisher$ and the Generalized Gauss-Newton Hessian $\ggn$ coincide, which we will use as a substitute for the Hessian $\mathbf{H}$. For details, see  \Cref{appendix:extended_preliminaries}.} so that its eigendecomposition can be done for each layer $l$ independently, which is significantly cheaper than an often infeasible brute-force eigendecomposition. To illustrate, for a $P$-layer multi-layer perceptron with $\tilde{D}$ input dimensions and $\tilde{D}$ output dimensions for all layers, the \ekfac \space eigendecomposition only costs $O(P\tilde{D}^3) = O(D^\frac{3}{2})$ time and storing its statistics requires $O(P\tilde{D}^2) = O(D)$ memory \citep{grosse2021nntd}.\footnote{The time and space complexity of ordinary eigendecomposition is $O(D^3)$ and $O(D^2)$ respectively, and $P\tilde{D}^3$ typically is much smaller than $D^3$. This analysis treats $P$ as constant.} Although \ekfac \space significantly reduces the time and space complexity of the eigendecomposition, it results in a biased \ihvp. See \Cref{appendix:extended_preliminaries} for a detailed discussion of \ekfac.

In contrast, iterative methods such as \lissa \space 
are based on Neumann series iterations (\vni) \citep{horn2012matrix}. \vni \space do not require 
\ekfac's assumptions, and thus in principle can produce exact \ihvps \space as more compute is applied. 
For an invertible matrix $\mathbf{A} \in \mathbb{R}^{D \times D}$ with $\|\mathbf{I} - \mathbf{A}\|_2 < 1$, the Neumann series is defined as
    $\mathbf{A}^{-1} = \sum_{j=0}^{\infty} \left( \mathbf{I} - \mathbf{A} \right)^j$,
which is a generalization of the geometric series $a^{-1} = \sum_{j=0}^{\infty} (1 - a)^j$ for $|1 - a| < 1$. By substituting $\mathbf{A}$ with a scaled positive-definite damped Generalized Gauss-Newton Hessian (GGN) $\alpha (\ggn + \lambda \eye)$, and multiplying both sides by any $\mathbf{v} \in \mathbb{R}^{D}$, we obtain:
$
    \frac{1}{\alpha} (\mathbf{G} + \lambda \mathbf{I})^{-1} \mathbf{v} = \sum_{j=0}^{\infty} \left( \mathbf{I} - \alpha \mathbf{G} - \alpha \lambda \mathbf{I} \right)^j \mathbf{v}
$. Here $\lambda$ is a positive scalar known as the damping term and $\alpha > 0$ is the learning rate hyperparameter. The learning rate must satisfy $\alpha < \frac{1}{\sigma_\text{max}(\mathbf{G}) + \lambda}$, where $\sigma_{\text{max}} (\mathbf{G})$ is the largest eigenvalue of $\mathbf{G}$ so that $\|\mathbf{I} - \mathbf{A}\|_2 < 1$. We can then approximate the \ihvp \space $\frac{1}{\alpha}(\mathbf{G} + \lambda \mathbf{I})^{-1} \mathbf{v}$ via the iterative update:
\begin{align}
    \mathbf{v}_{k+1} 
    \leftarrow \mathbf{v}_k - \alpha (\mathbf{G} + \lambda \eye) \mathbf{v}_{k} + \mathbf{v},
    \label{eqn:general_neumann_update}
\end{align}
which satisfies the property $\mathbf{v}_{k} \rightarrow \frac{1}{\alpha}(\mathbf{G} + \lambda \mathbf{I})^{-1} \mathbf{v}$ as $k \rightarrow \infty$. Computing  $\ggn$ requires two backward passes over the whole dataset,\footnote{An efficient implementation with $O(D)$ time and space complexity requires a Jacobian-vector product and a vector-Jacobian product \citep{pearlmutter1994fast}.} so an unbiased estimate  $\tilde{\ggn}_k$ of $\ggn$ is usually used instead by sampling a mini-batch with replacement, which we refer to as \textit{stochastic Neumann series iterations} (\sni). Compared to other iterative methods like conjugate gradient (\conjugategradient) \citep{hestenes1952methods}, \sni \space is typically preferred to compute the \ihvp \space for TDA since CG tends to struggle with stochastic gradients \citep{koh2017understanding, martens2015optimizing, snoek2012practical}. \lissa \space reduces the variance in \sni \space by taking an average over multiple trials.

\subsection{Training Data Attribution with Influence Functions}\label{influence_functions}
Influence functions \citep{hampel1974influence, cook1979influential} are derived under the assumption that $\mathcal{J}$ is strictly convex in $\params$ and twice differentiable. Let $\params^\star \coloneqq \argmin_{\params} \cost(\params, \mathcal{D})$ be the optimal parameters over $\mathcal{D}$. We can define the objective after downweighting a training example $\dataPoint_m$ by $\epsilon$ as: 
    $\mathcal{Q}(\params, \epsilon) \coloneqq \mathcal{J}(\params, \mathcal{D}) - \frac{\epsilon}{N} \mathcal{L} (\params, \dataPoint_m)$
 where $\epsilon$ is a scalar that specifies the amount of downweighting. When $\epsilon = 0$, this corresponds to the original objective $\mathcal{J}$. We can then define the optimal parameters \textit{after} downweighting as a function of $\epsilon$: $r(\epsilon) = \argmin_{\params \in \mathbb{R}^D} \mathcal{Q}(\params, \epsilon)$. When $\dataPoint_m \in \mathcal{D}$ and $\epsilon = 1$, this corresponds to the downweighted objective in which $\dataPoint_m$ is removed from $\mathcal{D}$. Typically, $\epsilon$ is assumed to be small, and we can approximate the leave-one-out (LOO) parameter change as $\oparam(\mathcal{D} \setminus \{\dataPoint_m\}) - \oparam (\mathcal{D}) \approx \dd{r}{\epsilon} \Bigr|_{\epsilon=0}$,
where $\dd{r}{\epsilon}\Bigr|_{\epsilon=0} = \frac{1}{N}\invhessian \traingrad$ and $\mathbf{H} \coloneqq \nabla^2_{\params} \mathcal{J} (\params^\star, \mathcal{D})$ denotes the Hessian of the training objective at the optimal parameters. To approximate the effect on the measurement function $f_{\dataPoint_q}$, we invoke the chain rule:
$f_{\dataPoint_q}(\oparam(\mathcal{D} \setminus \{\dataPoint_m\})) - f_{\dataPoint_q}(\oparam (\mathcal{D})) \approx  \frac{1}{N}\querygrad^\top \invhessian \traingrad \label{eqn:influence_chain_rule}$,
which also gives the first-order Taylor approximation of $f_{\dataPoint_q}(\oparam(\mathcal{D} \setminus \{\dataPoint_m\}))$ after rearranging terms. 
When applying this approximation to neural networks in which the convexity assumption does not hold, $\hessian$ may not be invertible, so $\hessian$ is typically approximated with the damped GGN $\ggn + \lambda \eye$ \citep{schraudolph2002fast}, which is always positive definite for $\lambda > 0$, and tends to work well in practice \citep{martens2020new, grosse2023studyinglargelanguagemodel, bae2024training, mlodozeniec2024influence}.\footnote{We will use the approximation $\mathbf{G} \approx \mathbf{H}$ throughout, and our use of the term \ihvp \space will generally refer to both the inverse Hessian-vector product and the inverse Gauss-Newton-Hessian-vector product.} 
With this substitution, the influence functions \textit{attribution score} is:
\begin{align}
     \tau_{\text{IF}}(\dataPoint_m, \dataPoint_q, \mathcal{D}) \coloneqq  \querygrad^\top (\mathbf{G} + \lambda \mathbf{I})^{-1} \traingrad. \label{eqn:influence_tau_definition}
\end{align}
When applying influence functions to neural networks, the model may not have fully converged, so we typically compute the gradients and the GGN in  \Cref{eqn:influence_tau_definition} with the final parameters $\finalParams$ instead of $\oparam$. We can also ensemble influence functions for better TDA performance, by training models with various seeds $\xi$ and averaging over $\tau_{\text{IF}}$ for each seed to get an ensembled score \citep{park2023trak, bae2024training} (details in \Cref{appendix:extended_preliminaries}). Ensembling for other TDA methods, such as unrolled differentiation, can be done analogously. 

The \ihvp \space in \Cref{eqn:influence_tau_definition} was originally computed with \lissa \space by \citet{koh2017understanding} and has the drawback that its iterative procedure must be carried out once for each vector in the \ihvp. Fortunately, it is frequently the case that $|\testdata| \ll |\trainingData|$ \citep{bae2022if, grosse2023studyinglargelanguagemodel, ko2024mirrored}, so by choosing $\mathbf{v} \coloneqq \querygradwithfinalparamsnorand$ and first computing $\querygradwithfinalparamsnorand^\top (\ggn + \lambda \mathbf{I})^{-1}$ in \Cref{eqn:influence_tau_definition},\footnote{This uses the fact that $\ggn$ is symmetric. $(\querygradwithfinalparamsnorand^\top (\ggn + \lambda \mathbf{I})^{-1})^\top = (\ggn + \lambda \mathbf{I})^{-1} \querygradwithfinalparamsnorand$.} we can reduce the number of iterative procedures to $|\testdata|$ for the influence function computation. Plugging these values into \Cref{eqn:general_neumann_update}, we can compute $\querygradwithfinalparamsnorand^\top (\ggn + \lambda \mathbf{I})^{-1}$ via the iterative update:\footnote{We emphasize that $\tilde{\ggn}_k$ and $\querygradwithfinalparamsnorand$ are computed using the final parameters $\params^s$. We use $\param_k$ to denote the iterates for \sni \space since it has the same dimensions as the model's parameters.}
\begin{align}
    \mathbf{\params}_{k+1} \leftarrow \mathbf{\params}_k - \alpha (\tilde{\mathbf{G}}_k + \lambda \mathbf{I}) \param_k + \alpha \mathbf{\nabla_{\params}}f_{\mathbf{z}_q}(\params^{s}). \label{eqn:sni_update_rule}
\end{align}
While $|\testdata|$ sounds like a large number of iterative procedures, in practice no $\testdata$ exists, and instead queries are run when the user wants to understand specific behavior pertaining to $\dataPoint_q$. Nevertheless, to get a good approximation of $\querygradwithfinalparamsnorand^\top (\ggn + \lambda \mathbf{I})^{-1}$ for any particular $\dataPoint_q$ may require \textit{thousands} of iterations \citep{koh2017understanding, grosse2023studyinglargelanguagemodel}, limiting its scalability. Furthermore, tuning the hyperparameter for \sni \space is difficult \citep{koh2017understanding,schioppa2022scaling, klochkov2024revisiting} due to the ill-conditioning and stochasticity of the gradients. Once the \ihvp \space is computed, its dot product with  $\nabla_{\params} \mathcal{L}(\params^{s}, \dataPoint_m)$ for every $\dataPoint_m$ in consideration is taken. If the goal is to simply compute the influence of $\dataPoint_m$ on $\dataPoint_q$ for \textit{given} $\dataPoint_m$ and $\dataPoint_q$, then the cost of the dot product is minimal. However, if the goal is to search for the most influential points in $\dataPoint_m \in \mathcal{D}$ on $\dataPoint_q$, then we must take the dot product with \textit{every} training example gradient in $\mathcal{D}$, which amounts to a backward pass over the entire training dataset and can be a substantial component of the cost of computing influence functions.\footnote{Procedures such as TF-IDF filtering exist to prune the potentially vast training dataset \citep{grosse2023studyinglargelanguagemodel}.} 
Arguably, the latter goal is more prevalent in settings such as data-centric model debugging and interpretability, where insight into model behavior is given by retrieving highly influential training samples~\citep{brennen2020people, hammoudeh2024training, pezeshkpour2022combining, nguyen2024towards, shankar2022operationalizing}.

\section{Introducing \ekfac-Accelerated Neumann Series Iterations for TDA}\label{section:method}

\paragraph{Preconditioning Stochastic Neumann Series Iterations with \ekfac}
The \sni \space update rule in \Cref{eqn:sni_update_rule} can be viewed as performing mini-batch gradient descent on the 
quadratic objective~\citep{saad2003iterative, agarwal2017second}:
\begin{align}
    h_{f_{\dataPoint_q}}(\mathbf{\params}) \coloneqq \frac{1}{2} \mathbf{\params}^\top (\mathbf{G} + \lambda \mathbf{I}) \mathbf{\params} - \mathbf{\params}^\top \querygradwithfinalparamsnorand.\label{eqn:quadratic_objective}
\end{align}
It is well-known that for a converged neural network, the curvature matrix in the objective in \Cref{eqn:quadratic_objective} is typically ill-conditioned \citep{sagun2016eigenvalues, ghorbani2019investigation}, which presents challenges for iterative methods. %
To improve the conditioning of $h_{f_{\dataPoint_q}}(\mathbf{\params})$, we introduce an algorithm which computes accurate \ihvps \space for use in TDA called \ekfac-\textbf{A}ccelerated Neumann \textbf{S}eries Iterations for \textbf{T}\textbf{R}aining Data \textbf{A}ttribution (\algoname) by applying preconditioning. The resulting 
update rule is:
\begin{align}
    \mathbf{\params}_{k+1} \leftarrow \mathbf{\params}_k - \alpha (\mathbf{P} + \tilde{\lambda} \mathbf{I})^{-1}(\tilde{\mathbf{G}}_k + \lambda \mathbf{I}) \params_k + \alpha (\mathbf{P} + \tilde{\lambda} \mathbf{I})^{-1}\querygradwithfinalparamsnorand. \label{eqn:quadratic_update_rule}
\end{align}

While a number of choices of preconditioners exist, we choose the GGN computed with \ekfac \space (i.e.,  $\preconditioner \coloneq \ggn_{\text{EKFAC}}$) for the following reasons: First, the computation cost of the \ekfac \space eigendecomposition is usually much cheaper than full matrix inversion, and its eigendecomposition statistics can be stored compactly and shared across all $|\testdata|$ optimization problems, since the value of $\ggn$ is the same across all objectives. Second, this choice has a close connection with \ekfac \space influence functions: \Cref{eqn:general_neumann_update} suggests initializing $\params_0$ as $\querygradwithfinalparamsnorand$, which is frequently done in public implementations \citep{koh2017understanding}. Observe that if we initialize $\param_0 \leftarrow \mathbf{0}$, choose $\tilde{\lambda} = \lambda$ and a learning rate $\alpha=1$, we arrive at the \ihvp \space which would be approximated by \ekfac \space after one step of \algoname, resulting in the same TDA prediction as \ekfacif.\footnote{Going forward, we therefore directly initialize \algoname \space with $\params_0 \leftarrow (\mathbf{P} + \tilde{\lambda} \mathbf{I})^{-1}\querygradwithfinalparamsnorand$.} We hypothesize that further training using the update rule in \Cref{eqn:quadratic_update_rule} will improve the \ihvp \space approximation, a claim we validate empirically in \Cref{section:experiments}. This update rule assumes using mini-batch gradient descent, but our formulation is compatible with other optimization algorithms as well.

\paragraph{Time Complexity of \algoname} Computing the update in \Cref{eqn:quadratic_update_rule} involves explicitly constructing neither $(\mathbf{P} + \tilde{\lambda} \mathbf{I})^{-1}$ nor $(\mathbf{G} + \lambda \mathbf{I})$.  Instead, we use Hessian-vector products \citep{pearlmutter1994fast} and first compute $(\tilde{\mathbf{G}}_k + \lambda \mathbf{I}) \params_k$. We can then compute  $(\mathbf{P} + \tilde{\lambda} \mathbf{I})^{-1}(\tilde{\mathbf{G}}_k + \lambda \mathbf{I}) \params_k$ using the \ekfac \space preconditioner $\mathbf{P} \coloneq \mathbf{G}_{\text{EKFAC}}$, so the \textit{incremental} time complexity of \textit{each iteration} in our algorithm is $O(\mathcal{B}D)$ where $\mathcal{B}$ is the mini-batch size used to sample $\tilde{\mathbf{G}}_k$. For both \ekfacif \space and \algonameif, computing $\mathbf{G}_{\text{EKFAC}}$ is necessary to compute the \ihvp, which only needs to be done once per model and can be shared among all queries. When we need to search the entire dataset for highly influential training examples, both methods also need to compute the dot product of the resulting \ihvp \space with the training example gradients over all $\dataPoint_m$ in consideration, as discussed in \Cref{influence_functions}. In our experiments, \algonameif \space required only a few hundred incremental iterations, so its iterative component is a relatively small additional cost per query compared to the total cost to compute \ekfacif.

\paragraph{Application to Unrolled Differentiation} In addition to influence functions, we also study the use of \algoname \space in the context of an unrolling-based TDA method. Influence functions make assumptions such as model convergence and unique optimal parameters in its derivation, which may not be satisfied in practice \citep{koh2017understanding, bae2022if}. In contrast, unrolled differentiation methods such as \source \space sidestep this limitation by differentiating through the training trajectory. Here, we only sketch our approach to apply \algoname \space to \source, deferring the full discussion of the \source \space derivation to \Cref{appendix:extended_preliminaries} and the details of \algonamesource \space to \Cref{appendix:astra_source}. \source \space approximates differentiating through the training trajectory by partitioning it into $L$ segments, assuming stationary and independent GGNs and gradients within each. Its approximation of the first order effect of downweighting a training example contains $L$ different finite series involving the GGN, each of which can be approximated with an \ihvp. Similarly to \algonameif, \algonamesource \space improves this approximation by repurposing the \ekfac \space decompositions as preconditioners, which would have been needed to be computed to implement \source \space anyways.

\section{Relationship to Existing Works}\label{section:related_works}
\paragraph{TDA Methods using \ihvps} Both influence functions and unrolled differentiation can be viewed as belonging to a family of gradient-based TDA methods (for a survey see \citep{hammoudeh2024training}), which both have a connection with \ihvps.
Many gradient-based TDA methods are variants of the influence functions method proposed by \citet{koh2017understanding} with aims to improve its computational cost \citep{grosse2023studyinglargelanguagemodel, schioppa2022scaling, park2023trak}, by using techniques such as \ekfac \space for \ihvp \space approximation \citep{grosse2023studyinglargelanguagemodel}, Arnoldi iterations \citep{schioppa2022scaling}, gradient projection \citep{park2023trak, choe2024your}, and rank-one updates \citep{kwon2023datainf}, instead of iterative algorithms such as \lissa \space \citep{agarwal2017second} or \conjugategradient \space \citep{hestenes1952methods, fletcher1964function}, since the former is expensive and hard to tune \citep{koh2017understanding, schioppa2022scaling, klochkov2024revisiting}, and the latter struggles with stochastic gradients \citep{martens2010deep}.
Unrolled differentiation addresses the key derivation assumptions underlying influence functions -- namely, the convexity of the training objective and the convergence of the final model parameters \citep{bae2022if}. Methods include SGD-influence \citep{hara2019data}, HYDRA \citep{chen2021hydra}, \source \space \citep{bae2024training}, DVEmb \citep{wang2024capturing} and MAGIC \citep{ilyas2025magic}, which all differentiate through the training trajectory, and only differ in their approximations. Rather than comparing the TDA algorithms themselves, our goal in this paper is to show that substantial TDA performance improvements can be achieved by using better \ihvp \space approximations. In some public comparisons, \lissa-based influence functions perform poorly \citep{park2023trak, kwon2023datainf, deng2024texttt}, sometimes even worse than dot products\footnote{This TDA method, sometimes called Hessian-free \citep{kwon2023datainf, li2024influence}, simply takes the dot product between the training gradient and the query gradient and is equivalent to influence functions if the damped GGN is set to the identity. } \citep{kwon2023datainf, li2024influence} and many have opted to instead use \ekfacif \space as their method or baseline of choice \citep{bae2024training, choe2024your, mlodozeniec2024influence}. In principle, \ekfacif \space is only an \textit{approximation} of what \lissa-based influence functions attempts to compute, since the latter is only constrained by solver error while the former makes assumptions on the structure of the curvature matrix. We show in this paper that indeed, accurately solving the \ihvp \space typically produces better TDA performance than the \ekfac \space solution, which is what our algorithm \algoname \space addresses.

\paragraph{\ihvps \space beyond TDA} \ihvps \space can also be found in higher-order optimization algorithms such as Newton's method~\citep{nocedal1999numerical}, quasi-Newton methods~\citep{fletcher1963rapidly, broyden1965class, liu1989limited, nocedal1980updating}, natural gradient descent~\citep{amari1998natural, martens2010deep, zhang2018noisy}, KFAC \citep{martens2015optimizing}, and Hessian-free optimization~\citep{martens2010deep, martens2011learning}, which computes Hessian-vector products iteratively with \conjugategradient \space \citep{hestenes1952methods, fletcher1964function}. Influence functions can also be cast as a bilevel optimization problem \citep{maclaurin2015gradient, vicol2022implicit, lorraine2020optimizing, bae2025beyond}, which can be solved via implicit differentiation or unrolled differentiation.  Since \ihvps \space show up frequently in machine learning, there is motivation to adapt and develop \ihvp \space computation techniques such as \sni \space and \ekfac. While these two methods are well-established and preconditioning is a well-established  technique in optimization \citep{amari1998natural, martens2015optimizing}, %
to the best of our knowledge, no prior TDA method has combined these methods to compute the \ihvp.
For extended related works, see \Cref{appendix:extended_related}.

\section{Performance Comparisons}\label{section:experiments}

This section aims to answer the following questions: 1) Do \algonameif \space and \algonamesource \space outperform their \ekfac-based counterparts? 2) Is \algoname \space substantially faster than vanilla \sni? To answer these questions, we run experiments in a number of settings. For regression tasks, we use the UCI datasets Concrete and Parkinsons \citep{dua2019uci} trained with a multi-layer perceptron (MLP). For classification tasks, we use \cifarten \space \citep{krizhevsky2009learning} trained with \resnetnine \space \citep{he2016deep}, MNIST \citep{lecun2010mnist} and \fashionmnist \citep{xiao2017fashion} trained with MLPs, and  \gpttwo \space \citep{radford2019language} fine-tuned with \wikitexttwo  
 \space \citep{merity2016pointer}. We also include a non-converged setting, \fashionmnist-N, introduced by \citet{bae2024training}, for which \source \space was specifically designed; in this setting, 30\% of the training examples were randomly labeled, and the model was trained for only three epochs to avoid overfitting \citep{bae2024training}. In addition to comparing against EKFAC-IF and EKFAC-SOURCE, we also compare against two popular TDA methods TracIn \citep{pruthi2020estimating} and TRAK \citep{park2023trak}.
 Details for all experiments can be found in \Cref{appendix:experiment_details}.

\begin{figure*}[t!]
    \centering
    \includegraphics[width=\textwidth]{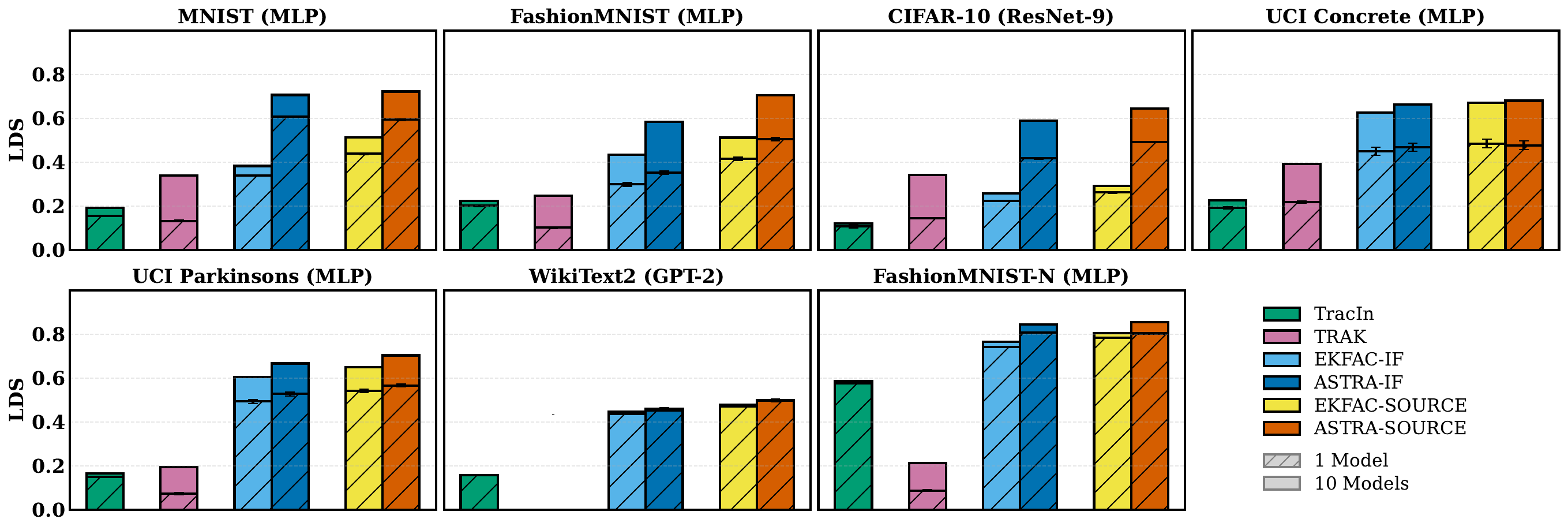}
    \captionsetup{font=small}
    \vspace{-1em}
    \caption{\small{\textbf{TDA Performance.} Single model \protect \algonameif \space and \algonamesource \space beat \protect \ekfac-based counterparts in most settings, as well as other TDA methods such as TracIn \citep{pruthi2020estimating} and TRAK \citep{park2023trak} when measured by average LDS over the query set $\testdata$. \protect \algoname \space also enjoys a larger performance boost from ensembling. Improvement is particularly large for convolution architectures such as \protect \resnetnine. Error bars (where available) indicate 1 standard error. We omit TRAK for \gpttwo \space due to lack of public implementations.}}
    \vspace{-0.5em}
    \label{fig:ensembling_benefits}
\end{figure*}
\begin{figure}[t!]
    \centering
    \includegraphics[width=\linewidth]{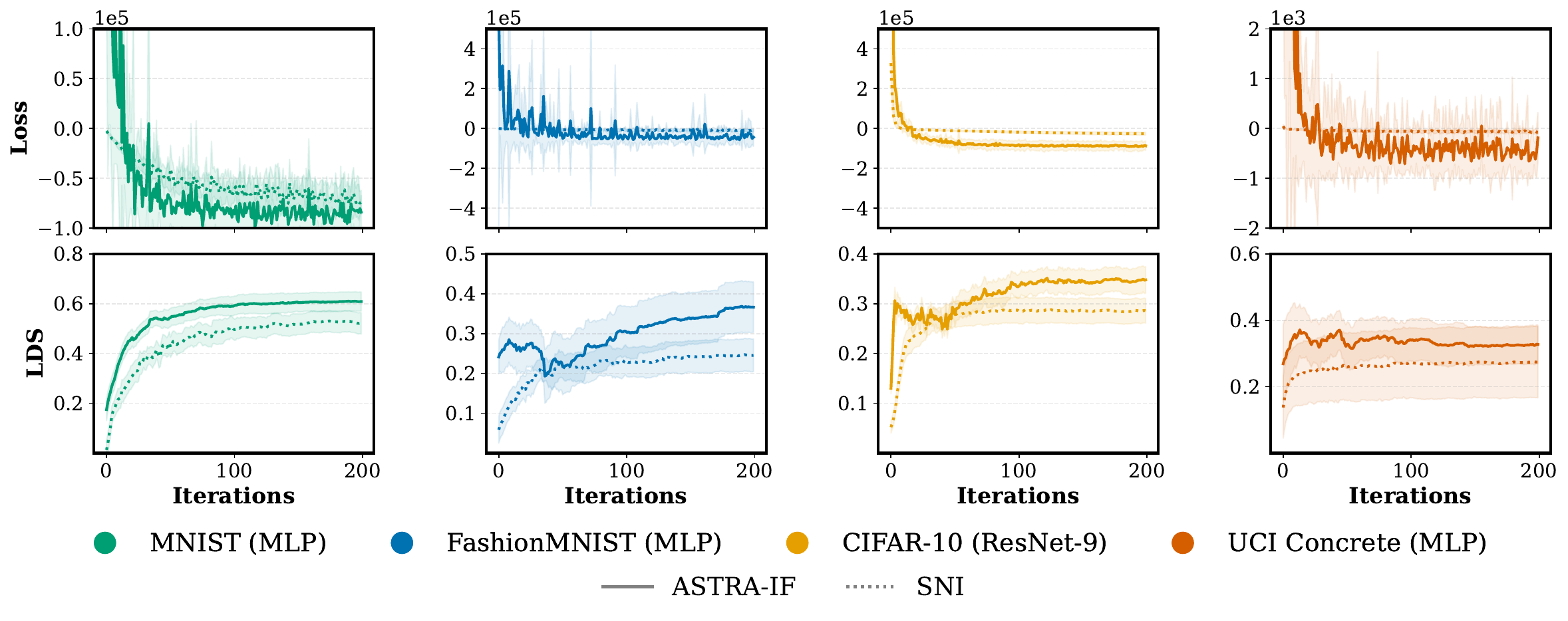}
    \captionsetup{font=small}
    \vspace{-1.2em}
    \caption{\small{\textbf{Training Curves}. Loss and LDS curves for \sni \space and \algoname \space measured over 10 seeds (shaded region = 1 standard error) on an arbitrary query point $\dataPoint_q$, using influence functions as the TDA method. \sni \space makes slower progress compared to \algoname \space as measured by LDS. }}
    \vspace{-0.5em}
    \label{fig:training_curves}
\end{figure}

\paragraph{Evaluating Training Data Attribution Performance}
We evaluate the performance of our TDA algorithms on a popular evaluation metric called Linear Datamodeling Score (LDS) \citep{park2023trak}, and use mean absolute error as the measurement function for regression tasks and correct-class margin for classification tasks in line with past works \citep{park2023trak, bae2024training}. LDS measures a TDA algorithm's ability to predict the outcome of counterfactual retraining on a subset of data. Given a collection of uniformly randomly sampled subsets $\{ \mathcal{S}_1, \ldots, \mathcal{S}_M: \mathcal{S}_i \subset \mathcal{D} \}$ of a fixed size, typically a fraction $\beta$ of $\mathcal{D}$ and a measurement function $f_{\dataPoint_q}$, the LDS scores a TDA method $\tau$ as follows:
\begin{align}
\text{LDS}(\tau, \dataPoint_q) &\coloneqq \boldsymbol{\rho}(\underbrace{\left[ \mathbb{E}_{\xi}[f_{\dataPoint_q} (\finalParams(\mathcal{S}_j; \xi))] : j \in [M] \right]}_{\substack{\text{\textit{expected} ground-truth predictions on  $\dataPoint_q$ } \\ \text{using models trained on various $\mathcal{S}_j$}}}, \underbrace{\left[ \Gamma_\tau(\mathcal{S}_j, \dataPoint_q; \mathcal{D}) : j \in [M] \right]}_{\substack{\text{TDA method's predictions on  $\dataPoint_q$} \\ \text{for various $\mathcal{S}_j$}}}), \label{eqn:lds_definition}
\end{align}
where $\boldsymbol{\rho}$ denotes the Spearman correlation \citep{spearman1904proof}, and the group influence $\Gamma_\tau( \mathcal{S}_j, \dataPoint_q, \mathcal{D})$ is defined linearly as:
$\Gamma_\tau(\mathcal{S}_j, \dataPoint_q; \mathcal{D}) = \sum_{z_i \in {\mathcal{S}_j}} \tau(\dataPoint_i, \dataPoint_q, \mathcal{D}).\label{eqn:lds_subdefinition}
$
To compute the ground-truth to which we compare our TDA method, we need to retrain a model many times both over various subsets $S_i$, and also over random seeds $\xi$ for every subset to obtain a good estimate of the expectation in \Cref{eqn:lds_definition}, which can be quite noisy \citep{nguyen2024bayesian, bae2024training}. For example, for the experiment involving \gpttwo, computing ground-truth involved fine-tuning 1000 models. We report the \textit{average} LDS  $\frac{1}{|\testdata|}\sum_{\dataPoint_q \in \testdata}\text{LDS}(\tau, \dataPoint_q)$ over a test set $\testdata$ containing 100 query points. We randomly sample $M=100$ subsets with a subsampling fraction of $\beta = 0.5$ in line with previous works \citep{bae2024training, choe2024your}. We discuss other evaluation methods in \Cref{appendix:evaluating_tda}.

\paragraph{LDS evaluation of \algoname}
\Cref{fig:ensembling_benefits} compares LDS across various TDA methods. We compare \algonameif \space and \algonamesource \space with their respective \ekfac-versions. For each setting, \ekfacif \space and \algonameif \space use the same damping value implied by \source \space for comparability (details in \Cref{appendix:experiment_details}). In almost all settings, \algoname \space improves TDA performance as measured by LDS, strongly suggesting that better \ihvp \space approximations are responsible. We also observe an especially large improvement over \ekfac \space for \cifarten \space trained on \resnetnine \space \citep{he2016deep}. When applied to convolution layers, \ekfac \space makes additional simplifying assumptions,\footnote{In addition to layer-wise independence and independence of activations and pseudogradients \citep{martens2015optimizing}, it assumes spatially uncorrelated derivatives and spatial homogeneity \citep{grosse2016kronecker}.} which can cause \ekfacif \space and \ekfacsource \space to underperform on architectures involving convolution layers -- an issue that \algoname \space effectively addresses. \Cref{fig:ensembling_benefits} also reveals increased benefits from ensembling when applying \algoname, which computes an unbiased estimator of the \ihvp. In some cases, such as \fashionmnist, the advantages of using precise \ihvps \space become much more pronounced in conjunction with ensembling. We hypothesize that this is due to the various bottlenecks in TDA methods: in some cases, the primary performance bottleneck lies in computing the iHVP accurately; in others, it stems from other factors such as the method's underlying assumptions (e.g., unique optimal parameters), which can be mitigated by ensembling.

\paragraph{\algoname \space speeds up \ihvp \space approximation}\label{subsection:computation}
The top row of \Cref{fig:training_curves} shows the loss curves for \sni \space and \algoname \space as each \ihvp \space solver progresses. The bottom row corresponds to the LDS that influence functions achieves based on the current progress of each \ihvp \space solver. For \sni, we follow public implementations \citep{koh2017understanding}, which typically initialize \sni \space using the query gradient $\querygradwithfinalparamsnorand$, and results in an initial influence functions prediction that approximates the damped-GGN with the identity in \Cref{eqn:influence_tau_definition}. For both methods, we conduct a hyperparameter sweep for the learning rate over $10^{0}$, \ldots , $10^{-5}$ in steps of one order of magnitude, and use the best hyperparameter based on the average training loss performance on the same query point over the last 10 iterations. We find that \ekfac \space preconditioning reduces the notorious challenge of tuning learning rates \citep{koh2017understanding, schioppa2022scaling, klochkov2024revisiting} -- in all of the settings reported in \Cref{fig:training_curves} the learning rate for \algoname \space used was $10^{-2}$. In comparison, the reported (and best) SNI learning rates were $1, 0.1, 0.01, 0.1$ for MNIST, FashionMNIST, CIFAR-10, and UCI Concrete respectively and LDS performance was very sensitive to the learning rate.  \Cref{fig:training_curves} shows that \sni \space makes slow progress while \algoname \space usually converges in fewer than 200 iterations.

\section{Investigating the Role of Low Curvature Directions in Influence Functions Performance}\label{section:truncated_neumann_series}

Our results in the previous section lead us to hypothesize that preconditioning accelerates convergence in directions of low curvature, which is important for influence function performance. We can analyze how directions of low curvature are affected by \vni \space (without preconditioning) when truncated early, something that is tempting to do as it usually takes long to converge. We derive the following expression for the truncated Neumann series with $J$ iterations: 
\begin{align}
\alpha \sum_{j=0}^{J-1} \left( \mathbf{I} - \alpha \ggn - \alpha \lambda \mathbf{I} \right)^j &= (\ggn + \lambda \mathbf{I})^{-1} (\mathbf{I} - (\mathbf{I} - \alpha \ggn - \alpha \lambda \mathbf{I})^J) \\
&\approx (\ggn + (\lambda + \alpha^{-1} J^{-1})  \mathbf{I})^{-1},
\end{align}
where the equality utilizes the definition for finite series, and the approximation is identical to that used in \citep{bae2024training} %
(see \Cref{appendix:truncated_neumann_series} for the derivation). From the last equation, we can see that truncating Neumann series effectively adds an \textit{implicit} damping term of $\sfrac{1}{\alpha J}$, which disproportionately affects directions of low curvature.

This insight prompts an investigation into the role of low‑curvature directions in influence functions performance since, in addition to the implicit damping effect mentioned above, some methods may discard them when projecting gradients into lower‑dimensional subspaces \citep{schioppa2022scaling, choe2024your}. Let  $\ggn_{\text{\ekfac}} = \hat{\mathbf{Q}}\hat{\mathbf{D}}\hat{\mathbf{Q}}^\top$ be the \ekfac \space eigendecomposition of the GGN at the final parameters. We study the importance of directions of varying curvature by doing the following: We bin the $D$ eigenvalues given by $\ggn_{\text{\ekfac}}$ into $5$ bins labeled $S_1, S_2, \ldots, S_5$, where each bin $S_i$ holds all eigenvalues larger than $10^{-i}$. Let $\hat{\mathbf{Q}}_{S_i} \in \mathbf{R}^{D \times |S_i|}$ be the projection matrix whose columns are the associated orthonormal eigenvectors of the eigenvalues in each bin.
We investigate the values of $h_{f_{\dataPoint_q}}^{S_i}(\mathbf{\params}_k)$ and the corresponding LDS for each $S_i$ where $h_{f_{\dataPoint_q}}^{S_i}$ is defined as:
\begin{align}
    h_{f_{\dataPoint_q}}^{S_i}(\mathbf{\params}) \coloneqq \frac{1}{2} (\hat{\mathbf{Q}}^\top_{S_i}\mathbf{\params})^\top \hat{\mathbf{Q}}^\top_{S_i}(\ggn + \lambda \mathbf{I})\hat{\mathbf{Q}}_{S_i}(\hat{\mathbf{Q}}_{S_i} ^\top\mathbf{\params})  - (\hat{\mathbf{Q}}_{S_i}^\top\mathbf{\params})^\top \hat{\mathbf{Q}}_{S_i}^\top \querygradwithfinalparamsnorand.\label{eqn:quadratic_objective_decomposed}
\end{align}
We use the \ekfac \space eigendecomposition since the true eigendecomposition is intractable for the settings we report. We conduct the experiment in the \mnist \space and \fashionmnist \space settings and use a small damping hyperparameter of $\lambda = 10^{-4}$ to be able to observe the impact of directions of low curvature on influence functions performance (details in \Cref{appendix:experiment_details}).

\begin{figure}[t!]
    \centering
    \includegraphics[width=\linewidth]{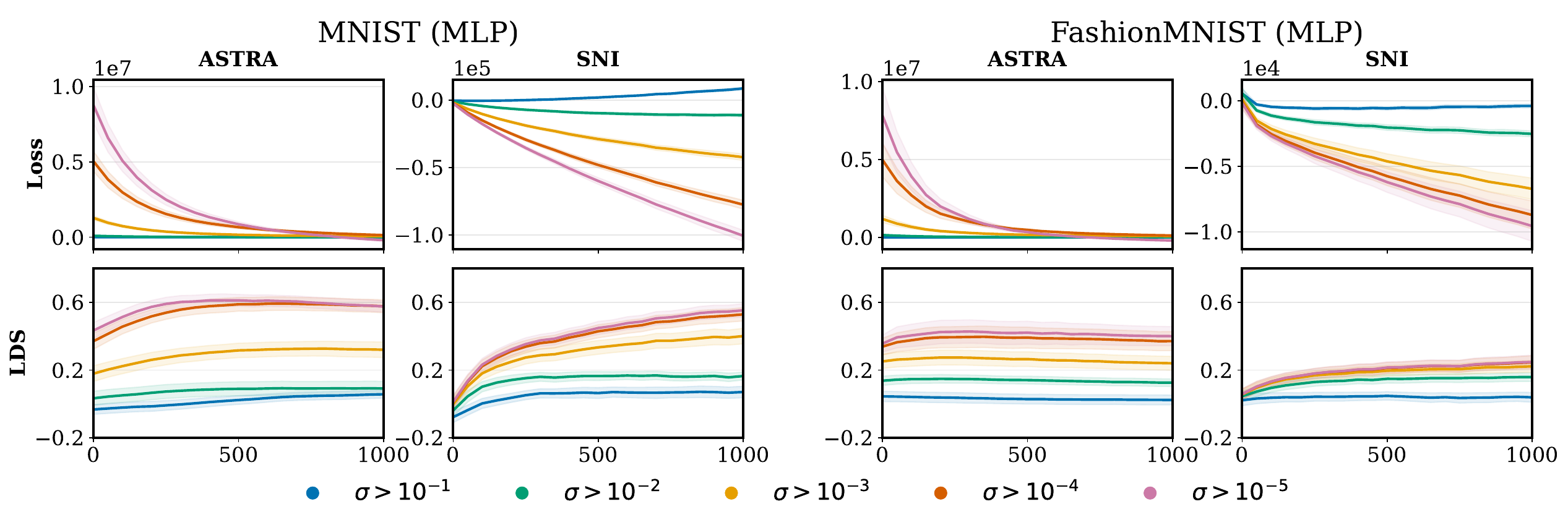}
    \captionsetup{font=small}
    \caption{\small{\textbf{Training Curves in Various Eigenspaces}. \textbf{Top:} The values $h_{f_{\dataPoint_q}}^{S_i}$ as \algoname \space and SNI train on the objective in \Cref{eqn:quadratic_objective} for an arbitrary $\dataPoint_q$. The subspaces represented by $S_1, \ldots, S_5$ are spanned by eigenvectors with eigenvalues $\sigma > 10^{-1}$, \ldots, $\sigma > 10^{-5}$ respectively. The loss of subspaces with large eigenvalue directions tend to plateau first, followed by subspaces with smaller eigenvalue directions, especially for \sni, which does not use preconditioning. \textbf{Bottom:} LDS of influence functions after projecting to corresponding subspace. The objective for \textit{high} curvature directions plateaus first; continued training further decreases the objective in progressively lower curvature directions, yielding LDS gains even after high curvature directions have converged. Shaded region = 1 standard error. }}
    \vspace{-0.8em}
    \label{fig:decomposition}
\end{figure}

The top row of \Cref{fig:decomposition} shows the outcome when we run \algoname \space and \sni \space on the objective in \Cref{eqn:quadratic_objective} to obtain a sequence of $\params_k$, and plot the value of each $h_{f_{\dataPoint_q}}^{S_i}(\params_k)$ for each $S_i$. The bottom row shows the LDS, which are computed by projecting the iterates $\params_k$ into each curvature subspace defined by $S_i$ and computing influence functions using these projected vectors, which we can write as:
\begin{align}
    \tau_{\text{PROJ-IF}, k}(\dataPoint_m, \dataPoint_q, \mathcal{D}) \coloneqq  (\hat{\mathbf{Q}}^\top_{S_i} \params_k)^\top \hat{\mathbf{Q}}^\top_{S_i} \traingradwithfinalparamsnorand.
\end{align}
\Cref{fig:decomposition} reveals that low curvature directions play a large role in the performance of influence functions: projecting to high-curvature eigenspaces degrades LDS performance, as evidenced by the large vertical gaps between lines representing different level of curvature. When the \ihvp \space is computed via \sni, large eigenvalue directions converge quickly during training, but it takes longer for small eigenvalue directions to converge, as evidenced by the earlier plateau of the loss curves in the high‑curvature subspaces in the top row of \Cref{fig:decomposition}. Nevertheless, as the solution progresses in low‑curvature directions, LDS rises substantially, evidenced by the growing gap between lines representing large and small levels of curvature in the bottom row of \Cref{fig:decomposition}. While the slower convergence in low-curvature directions is present for \algoname, it is substantially diminished due to the preconditioning.
Our results highlight that the behavior of estimators in low-curvature subspaces may be a substantial factor in the performance of TDA methods.\footnote{We note that the difference in performance between \algoname \space and \sni \space depends on the damping hyperparameter, and since the damping in this experiment is smaller, the performance boost from \algoname \space compared to SNI is notably larger than in \Cref{fig:training_curves}.}

\section{Conclusion}\label{conclusion}
We presented an algorithm \algoname \space that combines the \ekfac \space preconditioner with \sni \space for TDA. We compared \algonameif \space and \algonamesource \space with their \ekfac-based counterparts in a variety of settings. In many settings, TDA performance measured by LDS improved substantially, especially for convolution architectures. We find that in general, a more accurate \ihvp \space approximation increases the efficacy of ensembling. \algoname \space is easier to tune and converges faster than \sni. Compared with \ekfac, it only incrementally costs hundreds of iterations in our experiments as it leverages the same eigendecomposition. We conclude this paper by providing insights into how various curvature directions affect influence functions performance. By leveraging the \ekfac \space eigendecomposition, we show that low eigenvalue directions are important for good influence functions performance. We discuss limitations and broader implications of our work in \Cref{appendix:limitations_and_broader_impact}.

\section*{Acknowledgements}
We gratefully acknowledge funding from the Natural Sciences and
Engineering Research Council of Canada (NSERC) and the Canada CIFAR AI Chairs
Program. Resources used in preparing this research
were provided, in part, by the Province of Ontario, the Government of
Canada through CIFAR, and companies sponsoring the Vector Institute for
Artificial Intelligence (\url{https://vectorinstitute.ai/partnerships/}). We thank the Schwartz Reisman Institute for Technology and Society for
providing a rich multi-disciplinary research environment. RG and SM acknowledge support from the Canada CIFAR AI Chairs program and from the Natural Sciences and Engineering Research Council of Canada (NSERC).

\bibliographystyle{unsrtnat} 
\bibliography{bibliography}

\ifneurips
    \input{neurips_checklist}
\fi

\appendix
\pagebreak

\raggedbottom

\section{Notation \& Acronyms}\label{appendix:notation}

\subsection{Notation}
\begin{table}[!h]
    \small
    \begin{center}
        \begin{tabular}{c  p{0.81\textwidth}}
            \textbf{Notation}         & \textbf{Description} \\
            \midrule
            
            $B$ & Batch size in mini-batch gradient descent\\
            $\mathcal{B}$ & Batch size for stochastic Neumann series iterations \\
            $D$ & Number of parameters in neural network\\
            $J$ & Number of iterations for \sni\\
            $M$ & Number of subsets (masks) to sample for LDS computation\\
            $N$ & Number of training data points, $N = |\trainingData|$\\
            $P$ & Number of layers in a neural network.\\
            $R$ & Number of trials (repeats) for \lissa.\\
            \midrule
            $\mathcal{D} = \{\dataPoint_i\}_{i=1}^N$ & Training dataset \\
            $\testdata$ & Query dataset, used to benchmark TDA algorithms and small in practice \\
            $\mathcal{S}$ & Data subset of the training dataset \\
            $\dataPoint_i$ & An arbitrary $i$-th training example\\
            $\dataPoint_m \in \mathcal{D}$ & A training example from the dataset $\mathcal{D}$ \\
            $\dataPoint_q$ & A query data point\\
            $\inp^{(i)}$ & The inputs (feature vector) of the $i$‑th training example\\
            $\mathbf{z}^{(i)}$ & The neural network output for the $i$‑th training example\\
            $\target^{(i)}$ & The ground‑truth target for the $i$‑th example \\
            $\hat{\mathbf{y}}^{(i)}$ & Sampled target for the $i$‑th example using model probabilities\\
            \midrule
            $\xi$ & Source of training procedure randomness\\
            $\xi_b$ & Randomness from batch ordering\\
            $\Xi$ & A set containing various seeds $\xi$\\
            $\params$ & The neural network parameters\\
            $\optParams$ & Optimal parameters\\
            $\optParams(\mathcal{S})$ & Optimal parameters trained on data subset $\mathcal{S} \subseteq \mathcal{D}$\\
            $\finalParams$ & Final model parameters (not necessarily at optimum)\\
            $\finalParams(\xi)$ & Final model parameters (not necessarily at optimum) which depends on randomness $\xi$\\
            $\params_k$ & The parameters of a network after $k$ iterations of an algorithm, which depends on context\\
            \midrule
            $g(\params, \inp)$ & The output (logits) of a neural network with parameters $\params$ and input $\inp$\\
            $\loss(\mathbf{z}, \mathbf{t})$ & Loss function as a function of the neural network output and target (e.g.,  cross-entropy)\\
            $\loss(\params, \dataPoint)$ & Loss function as a function of the parameters and training example\\
            $\cost(\params, \trainingData)$ & Cost function, $\cost(\params, \trainingData) = \frac{1}{N} \sum_{i=1}^N \loss (\params, \dataPoint_i)$ \\
            $f_{\dataPoint_q}(\params)$ & Measurement function on query point $\dataPoint_q$, typically correct-class margin or absolute error\\
            $h_{f_{\dataPoint_q}}(\mathbf{\params})$ & The Neumann series iteration objective for the \ihvp \space $\querygradwithfinalparamsnorand^\top (\ggn + \lambda \mathbf{I})^{-1}$\\
            $h_{f_{\dataPoint_q}}^{S_i}(\params)$ & The Neumann series iteration objective projected onto the subspace corresponding to $S_i$\\
            \midrule
            $\mathbf{F}$ & The Fisher information matrix\\
            $\ggn$ & The Generalized Gauss-Newton Hessian (GGN) matrix\\
            $\ggn_{\text{\ekfac}}$ & The GGN matrix approximated with \ekfac \\
            $\tilde{\mathbf{G}}_k$ & An unbiased sample of the GGN computed with data at iteration $k$\\
            $\mathbf{H}$ & Hessian matrix $\mathbf{H} \coloneqq \nabla^2_{\params} \mathcal{J} (\params^\star, \mathcal{D})$\\
            $\mathbf{P}$ & Preconditioning matrix used in \algoname, chosen as the $\ggn_{\text{\ekfac}}$ \\
            \midrule
            $\beta$ & Fraction of $\trainingData$ used for subsampling when computing LDS\\
            $\eta$ & Learning rate when training the neural network\\
            $\alpha$ & Learning rate to find the \ihvp \space with \sni \space or \algoname\\
            $\epsilon$ & Downweighting amount used in influence functions and SOURCE formulation\\
            $\lambda$ & Damping parameter to compute \ihvps: a small positive scalar. \\
            $\tilde{\lambda}$ & Damping parameter added to the preconditioner:  a small positive scalar\\
            $\tau$ & Training data attribution method \\
            $\sigma$ & Eigenvalues in the decomposition of the curvature matrix. \\
            $\Gamma$ & Group influence of a training data attribution method \\
            $\boldsymbol{\rho}$ & The Spearman correlation coefficient \citep{spearman1904proof}\\
            $\otimes$ & The Kronecker product\\
            \bottomrule
        \end{tabular}
    \end{center}
    \vspace{-0.03\textheight}
    \label{tab:table-of-notation}
\end{table}
\pagebreak
\subsection{SOURCE specific notation}
\begin{table}[!h]
    \begin{center}
        \begin{tabular}{c  l}
            \textbf{Notation}         & \textbf{Description} \\
            \midrule
            $\delta_{ki}$ & Indicator variable used in SOURCE formulation\\
            ${\ell}$ & An index variable indicating the current segment in SOURCE\\
            $K_{\ell}$ & The number of iterations within segment $\ell$ in SOURCE\\
            $L$ & The number of segments in SOURCE\\
            $T$ & The number of optimization steps for the underlying model\\
            $\overline{\mathbf{g}}_{\ell}$ & The average gradient in segment $\ell$\\
            $\overline{\eta}_{\ell}$ & The average learning rate in segment $\ell$\\
            $\overline{\mathbf{r}}_{\ell}$ & Defined in \Cref{eqn:source_approximations_final}\\
            $\overline{\mathbf{S}}_{\ell}$ & Defined in \Cref{eqn:source_approximations_final}\\
            $\tilde{\mathbf{r}}_{\ell}$ & An approximation of $\overline{\mathbf{r}}_{\ell}$ introduced by \citet{bae2024training}\\
            \bottomrule
        \end{tabular}
    \end{center}
    \vspace{-0.03\textheight}
    \label{tab:table-of-source-notation}
\end{table}

\bigskip

\subsection{Acronyms}
\begin{table}[!h]
    \begin{center}
        \begin{tabular}{c  l}
            \textbf{Acronym}         & \textbf{Description} \\
            \midrule
            \conjugategradient & Conjugate Gradient \citep{hestenes1952methods}\\
            \ekfac & Eigenvalue-corrected Kronecker-factored Approximate Curvature \citep{george2018fast} \\
            FIM & Fisher Information Matrix \\
            \ihvp & Inverse Hessian-vector product, or inverse Gauss-Newton-Hessian-vector product\\
            IF & Influence functions \citep{koh2017understanding} \\
            KFAC & Kronecker-factored Approximate Curvature \citep{martens2015optimizing} \\
            \lissa &  Linear time Stochastic Second-Order Algorithm \citep{agarwal2017second}\\
            LOO & Leave-one-out \\
            LDS & Linear Datamodeling Score \citep{park2023trak} \\
            MLP & Multi-layer perceptron \\
            NI & Neumann series iterations\\
            PBRF & Proximal-Bregman Response Function \citep{bae2022if}\\
            SGD & Stochastic gradient descent\\
            SOURCE & Segmented statiOnary UnRolling for Counterfactual Estimation \citep{bae2024training}\\
            SNI & Stochastic Neumann series iterations\\
            TDA & Training data attribution \\
            \bottomrule
        \end{tabular}
    \end{center}
    \vspace{-0.03\textheight}
    \label{tab:table-of-acronyms}
\end{table}

\clearpage
\section{Extended Preliminaries}\label{appendix:extended_preliminaries}

\subsection{\lissa \space and \sni}\label{appendix:influence_derivation}

\lissa \space \citep{agarwal2017second} is a second-order optimization algorithm which involves the computation of an \ihvp. \citet{koh2017understanding} choose \lissa \space as their \ihvp \space solver, but \lissa \space contains other components as well. In this paper, ``\lissa'' refers to the \ihvp \space component. \Cref{alg:lissa} shows that the primary difference between \lissa \space and \sni \space is that the former repeats the \sni \space procedure multiple times to reduce variance (highlighted in {\color{red} red}). We use $R=1$ throughout this paper, so \lissa's \ihvp \space component is equivalent to \sni \space and thus we use the two terms interchangeably.
\begin{algorithm}[t]
\begin{algorithmic}
\Require $\mathbf{v} \in \mathbb{R}^D$, $\alpha > 0$ (learning rate), $J > 0$ (number of iterations), $R > 0$ (repeat size), $\lambda > 0$ (damping term), $\mathcal{B} > 0$ (batch size), $\mathcal{D}$ (training dataset)
\State $\mathbf{x} \gets \mathbf{0}$ \Comment{Initialize the accumulator for final estimation}
\For{$r = 1$ {\color{red} to $R$}}
\State $\mathbf{v}_0 \gets \mathbf{v}$ \Comment{Initialize $\mathbf{v}_0$ as per the initial condition}
\For{$j = 0$ to $J-1$}
    \State $\boldsymbol{\mathcal{B}} \gets \text{SampleWithReplacement}(\mathcal{D}, \mathcal{B})$ \Comment{Sample a mini-batch of size $\mathcal{B}$ from $\mathcal{D}$}
    \State $\mathbf{p} \gets \tilde{\mathbf{\ggn}}_{\boldsymbol{\mathcal{B}}} \mathbf{v}_k$ \Comment{Compute HVP using mini-batch $\mathcal{B}$.}
    \State $\mathbf{v}_{k+1} \gets \mathbf{v}_k - \alpha (\mathbf{p} + \lambda \mathbf{v}_k) + \alpha \mathbf{v}$ \Comment{\sni \space update rule}
\EndFor
\State $\mathbf{x} \gets \mathbf{x} + \mathbf{v}_K$ \Comment{Accumulate the result of this repetition}
\EndFor
\State  $\mathbf{x} \gets \mathbf{x} {\color{red} / R}$ \Comment{Average the accumulated results over $R$ repetitions}\\
\Return $\mathbf{x}$ \Comment{Return final iHVP estimation $\mathbf{H}^{-1} \mathbf{v}$}
\end{algorithmic}
\caption{iHVP approximation with \protect \lissa}
\label{alg:lissa}
\end{algorithm}

\subsection{Influence Functions}\label{appendix:influence_derivation}
Previous papers have shown that influence function estimates are often fragile due to the strong assumptions in the influence function derivation \citep{basu2020influence, bae2022if, epifano2023revisiting, sogaard2021revisiting, schioppa2024theoretical, hu2024most}. In table \Cref{tab:influence_assumptions}, we outline the main assumptions.

\begin{table}[h]
    \centering
    \captionsetup{font=small}
    \caption{\small{Summary of assumptions of Influence Functions vs. Unrolled Differentiation.}}
    \small
    \label{tab:influence_assumptions}
    \begin{tabular}{lp{0.15\linewidth}p{0.15\linewidth}}
        \toprule
         \textbf{Assumption} & \textbf{Influence \allowbreak Functions} & \textbf{Unrolled \allowbreak Differentiation} \\
         \midrule
         First order approximation & \cmark & \cmark \\
         Objective convex with respect to parameters  & \cmark & \xmark \\
         Model trained to optimal parameters & \cmark & \xmark \\
         \bottomrule
    \end{tabular}
\end{table}

Despite the strong assumptions in the derivation of influence functions, a poor \ihvp \space approximation can make influence functions estimates appear less reliable than they are. To appreciate these assumptions, we refer the reader to Appendix B.1 of \cite{bae2022if} for a well-presented derivation of influence functions.

\paragraph{Ensembling Influence Functions} TDA scores can be typically ensembled over multiple training trajectories \citep{park2023trak, bae2024training} to mitigate the problem of noise in the training procedure \citep{epifano2023revisiting, nguyen2024bayesian}. This is typically done by training models with various seeds $\xi \in \Xi$, and approximating the \textit{expected} first-order downweighting effect with the empirical average of attribution scores $\tau$:
\begin{align}
     \tau_{\text{IF-Ensemble}}(\dataPoint_m, \dataPoint_q, \mathcal{D}) \coloneqq  \frac{1}{|\Xi|}\sum_{\xi \in \Xi} \querygradwithfinalparams^\top (\mathbf{G}_\xi + \lambda \mathbf{I})^{-1} \traingradwithfinalparams , \label{eqn:influence_ensemble_definition}
\end{align}
where $\ggn_\xi$ is the GGN computed at $\finalParams(\xi)$. We perform ensembling in this paper using the procedure in \Cref{eqn:influence_ensemble_definition}, and apply ensembling for \source \space analogously. Note that \Cref{eqn:influence_ensemble_definition} ensembles the attribution scores \citep{park2023trak, bae2024training}, while some other works ensemble the weights \citep{ng2024measuring, bae2024training}.

\subsection{Curvature Matrices}
We explain the relationships between curvature matrices below for completeness, which is heavily based on \citet{grosse2021nntd} and \citet{martens2020new}.
\paragraph {Approximating $\mathbf{H}$ with $\ggn$} Throughout this paper, we use the approximation $\ggn \approx \mathbf{H}$. Letting $\mathbf{z} = g(\params, \mathbf{x})$ denote the neural network output\footnote{Note the difference between the bold font $\mathbf{z}$, which refers to neural network outputs, with italicized font $\dataPoint$, which refers to a data point $\dataPoint = (\inp, \mathbf{t})$}, the GGN is equal to the Hessian if we drop the second term from the following decomposition of $\mathbf{H}$ \citep{grosse2021nntd}:
\begin{align}
\mathbf{H} = \frac{1}{N} \sum\limits_{(\inp^{(i)}, \mathbf{t}^{(i)}) \in \mathcal{D}}  [{\mathbf{J}_{\mathbf{z}^{(i)} \params}}^\top \mathbf{H}_{\mathbf{z}^{(i)}} \mathbf{J}_{\mathbf{z}^{(i)} \params} + \sum_j \pp{\mathcal{L}}{\mathbf{z}^{(i)}_j}\nabla^2_{\params}g(\params, \mathbf{x}^{(i)})_j],
\end{align}
where $\mathbf{J}_{\mathbf{z}^{(i)} \params}$ is the Jacobian matrix of the neural network's outputs with respect to the parameters for the $i$th training example, $\mathbf{H}_{\mathbf{z}^{(i)}} \coloneq \nabla_{\mathbf{z}}^{2} \mathcal{L}\bigl(\mathbf{z}^{(i)}, \mathbf{t}^{(i)} \bigr)$ refers to the Hessian of the loss function with respect to the neural network outputs for the $i$th training example, $\pp{\mathcal{L}}{\mathbf{z}^{(i)}_j}$ refers to the derivative of the loss function with respect to the $j$th neural network output for the $i$th training example and $\nabla^2_{\params}g(\params, \mathbf{x}^{(i)})_j$ refers to the Hessian of the $j$th neural network output for the $i$th training example with respect to the parameters. For a linear neural network, $\nabla^2_{\params}g(\params, \mathbf{x}^{(i)})_j = 0$, so the GGN is equal to the Hessian if we linearize the neural networks with respect to the parameters and only capture the curvature in the loss function. Linearization of the neural network is an approximation documented in previous works \citep{jacot2018neural, lee2019wide, wei2022more, martens2020new} and used frequently for influence functions if the damped GGN $\mathbf{G} + \lambda \eye$ is used \citep{bae2022if, park2023trak, grosse2023studyinglargelanguagemodel, mlodozeniec2024influence}.

\paragraph {Equivalence of $\mathbf{F}$ and $\ggn$} For the machine learning tasks that we consider, such as regression and classification tasks, the outputs of the neural network can be seen as the natural parameters of an exponential family. For these cases, the Fisher Information Matrix and the GGN coincide (i.e. $\mathbf{F} = \ggn$). We will illustrate this for softmax classification, but the case for regression can be derived similarly. The cross‐entropy loss for a training example $\dataPoint = (\inp, \mathbf{t})$  is $\mathcal{L}(\params, \dataPoint) = - \mathbf{t}^\top \log p_{\params}(\mathbf{y}| \inp)$ whose  gradient is:  $\nabla_{\params}\,\mathcal{L}(\params,\dataPoint) = \mathbf{J}_{\mathbf{z}\params}^{\top}\,\bigl(p_{\params}(\mathbf{y}| \inp) - \mathbf{t}\bigr)$. Then the following equalities hold:
\begin{align*}
\mathbf{F}
&= \frac{1}{N} \sum\limits_{(\inp^{(i)}, \mathbf{t}^{(i)}) \in \mathcal{D}} \mathbb{E}_{\hat{\mathbf{y}} \sim p_{\boldsymbol{\theta}}(\mathbf{y} | \inp^{(i)} )} \left[ \nabla_{\boldsymbol{\theta}} \log p_{\boldsymbol{\theta}} (\hat{\mathbf{y}} | \inp^{(i)} ) \nabla_{\boldsymbol{\theta}} \log p_{\boldsymbol{\theta}} (\hat{\mathbf{y}} | \inp^{(i)} )^\top \right] \\
&= \frac{1}{N} \sum\limits_{(\inp^{(i)}, \mathbf{t}^{(i)}) \in \mathcal{D}}
   \mathbb{E}_{\hat{\mathbf{y}}\sim p_{\params}(\mathbf{y}\mid \mathbf{x}^{(i)})}
   \Bigl[ \mathbf{J}_{\mathbf{z}^{(i)} \params}^{\top} (p_{\params}(\hat{\mathbf{y}}\mid \mathbf{x}^{(i)})- \hat{\mathbf{y}} ) (p_{\params}(\hat{\mathbf{y}}\mid \mathbf{x}^{(i)})-\hat{\mathbf{y}} )^{\top} \mathbf{J}_{\mathbf{z}^{(i)} \params} \Bigr]
\\
&= \frac{1}{N} \sum\limits_{(\inp^{(i)}, \mathbf{t}^{(i)}) \in \mathcal{D}}
   \mathbf{J}_{\mathbf{z}^{(i)} \params}^{\top}
   \Bigl(\mathrm{diag}\bigl(p_{\params}(\hat{\mathbf{y}} \mid \mathbf{x}^{(i)})\bigr)
         \;-\;p_{\params}(\hat{\mathbf{y}} \mid \mathbf{x}^{(i)})\,p_{\params}(\hat{\mathbf{y}} \mid \mathbf{x}^{(i)})^{\top}\Bigr)
   \mathbf{J}_{\mathbf{z}^{(i)} \params}
\\
&= \frac{1}{N} \sum\limits_{(\inp^{(i)}, \mathbf{t}^{(i)}) \in \mathcal{D}}
   \mathbf{J}_{\mathbf{z}^{(i)} \params}^{\!\top}
   \nabla_{\mathbf{z}}^{2} \mathcal{L}\bigl(\mathbf{z}^{(i)}, \mathbf{t}^{(i)} \bigr)
   \mathbf{J}_{\mathbf{z}^{(i)} \params} = \ggn
\end{align*}

\subsection{Training Data Attribution with Unrolled Differentiation}\label{appendix:unrolled}
Influence functions may struggle with models that have not converged \citep{pruthi2020estimating, bae2022if, schioppa2024theoretical, bae2024training} (\Cref{tab:influence_assumptions}). Fortunately, unrolled differentiation methods \citep{hara2019data, chen2021hydra, bae2024training, wang2024capturing, ilyas2025magic} do not rely on model convergence. Instead, they capture the effect of downweighting a training example by differentiating through the entire training trajectory. They can also capture the effect of other sources of randomness, such as batch ordering \citep{li2014efficient}. Assume our optimization algorithm is mini-batch gradient descent, which uses a learning rate $\eta_k$ and batch size $B$, and let $\delta_{ki}$ be an indicator variable that equals 1 if and only if $\dataPoint_m = \dataPoint_{ki}$, where $\dataPoint_{ki}$ is the $i$-th training example in batch $k$.\footnote{We note that $\params_k$ in this setting refers to the parameters at time step $k$ when training the network and distinguish it from \sni \space or \algoname \space iterations presented in \Cref{background} and \Cref{section:method}.} Then the mini-batch gradient descent update rule is:
\begin{align}
    \params_{k+1}(\epsilon) \leftarrow  \params_k(\epsilon) &- \frac{\LR_k}{B} \! \sum_{i=1}^B (1 \!-\! \delta_{ki} \epsilon) \nabla_{\params} \loss(\params_k(\epsilon), \dataPoint_{ki}),
\end{align}
where $\delta_k \coloneq \sum_{i=1}^B \delta_{ki}$. Let $\xi_{b}$ denote the randomness from batch ordering. Then the expected first-order effect on the  parameters $\params_T$ after $T$ steps of training with $\dataPoint_m$ downweighted by $\epsilon$ is:\footnote{Unless otherwise stated, derivatives taken with respect to $\epsilon$ are evaluated at $\epsilon = 0$. The notation $\prod_{i=T-1}^{k+1}$ means taking products in decreasing order from $T-1$ to $k+1$.}
\begin{align}
    \mathbb{E}_{\xi_b}\left[\frac{\mathrm{d} \params_{T}}{\mathrm{d} \epsilon}\right] 
    &= -\mathbb{E}_{\xi_b}\left[\sum_{k=0}^{T-1} \frac{\LR_k}{B}  \delta_k  \left( \prod_{i=T-1}^{k+1}(\mathbf{I} - \eta_i \ggn_i) \right) \traingradwithinputs{\params_k}{\dataPoint_m}\right],
    \label{eqn:unroll_response}
\end{align}
\citet{bae2024training} introduce an algorithm called \source, which approximates \Cref{eqn:unroll_response} much more cheaply by segmenting the trajectory into $L$ segments, assuming stationary and independent GGNs and gradients within each segment. For the $\ell$th segment which starts at iteration $T_{\ell -1}$ and ends at iteration $T_{\ell}$, and $k$ satisfying $T_{\ell-1} \leq k < T_{\ell}$, let $\overline{\ggn}_{\ell} \coloneq \mathbb{E}_{\xi_b}[\ggn_k]$, $\overline{\mathbf{g}}_{\ell} \coloneq \mathbb{E}_{\xi_b}[\traingradwithinputs{\params_k}{\dataPoint_m}] $, and $ \overline{\eta}_{\ell}$ refer to the average GGN, average gradient, and average learning rate in segment $\ell$ respectively and let $K_{\ell} \coloneq T_{\ell} - T_{\ell-1}$ refer to the total number of iterations in segment $\ell$. Then \source \space approximates the first-order effect of downweighting parameters as:
\begin{align}
    \E _{\xi_b} \left[ \frac{\mathrm{d} \params_{T}}{\mathrm{d} \epsilon} \right]
    &\approx - \frac{1}{N}\sum_{\ell=1}^L \left( \prod_{\ell'=L}^{\ell+1} (\eye - \overline{\eta}_{\ell'} \overline{\ggn}_{{\ell'}})^{K_{\ell'}} \right)\left( \sum\limits_{k=T_{\ell-1}}^{T_{\ell}-1} \overline{\LR}_{\ell} (\eye - \overline{\eta}_{\ell} \overline{\ggn}_{{\ell}})^{T_{\ell}-1-k} \overline{\mathbf{g}}_{\ell} \right)  \label{eqn:source_approximations}\\
    &\approx - \frac{1}{N} \sum_{\ell=1}^L \left( \prod_{\ell'=L}^{\ell+1} \underbrace{\exp(-\overline{\eta}_{\ell'}K_{\ell'}\overline{\ggn}_{\ell'})}_{\overline{\segment}_{\ell'}} \right) \underbrace{\left( \eye - \exp(- \overline{\eta}_{\ell}K_{\ell}\overline{\ggn}_{\ell}) \right)\overline{\ggn}_{\ell}^{-1} \overline{\mathbf{g}}_{\ell} }_{\overline{\mathbf{r}}_\ell}.\label{eqn:source_approximations_final} 
\end{align}
The stationary and independent assumptions allow us to factor shared products resulting in the approximation in \Cref{eqn:source_approximations}, which can be then approximated with matrix exponentials in \Cref{eqn:source_approximations_final} and computed with the \ekfac \space eigendecomposition. \citet{bae2024training} provide an interpretation of the $\overline{\mathbf{r}}_\ell$ term, noticing that $\overline{\mathbf{r}}_{\ell} \approx \left( \overline{\ggn}_{\ell} + 
 \overline{\eta}_{\ell}^{-1} K_{\ell}^{-1} \eye \right)^{-1} \overline{\mathbf{g}}_{\ell} \coloneq \tilde{\mathbf{r}}_{\ell}$, which is an \ihvp \space that we can use to apply \algoname. We discuss this term in more detail in \Cref{appendix:astra_source}. We refer the reader to \citet{bae2024training} for a full explanation of \source.

\subsection{Kronecker-Factored Approximate Curvature}
The KFAC approximation was introduced in \citep{martens2015optimizing} in the context of second-order optimization and explained in \citep{grosse2023studyinglargelanguagemodel} in the context of influence functions. To understand the cost of \ekfac \space and \ekfacif \space compared to \algonameif, as well as the assumptions \ekfac \space make which results in a biased \ihvp \space approximation, we present the derivation below which is heavily based on \citet{grosse2021nntd} and \citet{grosse2023studyinglargelanguagemodel} and refer readers to \citet{martens2015optimizing} and \citet{george2018fast} for further reading.

Our goal is to compute the iHVP with the Fisher Information Matrix (FIM) $\mathbf{F}$ as an approximation for the Hessian $\mathbf{H}$. The FIM is defined as:
\begin{align}
    \mathbf{F} &\coloneqq \frac{1}{N}\sum\limits_{\inp^{(i)} \in \mathcal{D}}\mathbb{E}_{\hat{\mathbf{y}} \sim p_{\boldsymbol{\theta}}(\mathbf{y} | \inp^{(i)})} \left[ \nabla_{\boldsymbol{\theta}} \log p_{\boldsymbol{\theta}} (\hat{\mathbf{y}} | \inp^{(i)}) \nabla_{\boldsymbol{\theta}} \log p_{\boldsymbol{\theta}} (\hat{\mathbf{y}} | \inp^{(i)})^\top \right] \label{eqn:fisher_expectation_definition}
\end{align}

\noindent
where $p_{\boldsymbol{\theta}}(\mathbf{y} | \mathbf{x})$ is the model's own distribution over targets. We omit the random variables in the expectation's subscripts going forward to reduce clutter. Using the model's own distribution over targets (as opposed to actual targets) is rather important since using the actual targets rather than the model's distribution results in a matrix called the Empirical Fisher, which does not have the same properties as the FIM \citep{kunstner2019limitations}.

We now describe KFAC for multilayer perceptrons. We refer readers to \citet{grosse2016kronecker} for the derivation for convolution layers. Consider the $l$-th layer of a neural network,\footnote{Note the difference between $\ell$, which refers to a segment in \source, and $l$, which is an index denoting a layer of a neural network.} which has input activations $\mathbf{a}_{l-1} \in \mathbb{R}^{I}$, weights $\mathbf{W}_{l} \in \mathbb{R}^{O \times I}$, bias $\mathbf{b}_{l} \in \mathbb{R}^{O}$, and outputs $\mathbf{s}_{l} \in \mathbb{R}^{O}$. For convenience, we use the notation $\overline{\mathbf{a}}_{l-1} = \bigl[\mathbf{a}_{l-1}^{\top}\;1\bigr]^{\top}$ and $\overline{\mathbf{W}}_{l} = [\mathbf{W}_{l}\;\mathbf{b}_{l}]$ to handle weights and biases together, and we write $\params_{l} = \mathrm{vec}(\overline{\mathbf{W}}_{l})$ to denote the reshaped vector of layer $l$ parameters. Then each layer computes:
\begin{equation}
\mathbf{s}_{l} \;=\; \overline{\mathbf{W}}_{l}\,\overline{\mathbf{a}}_{l-1},
\quad
\mathbf{a}_{l} \;=\; \phi_{l}(\mathbf{s}_{l}),
\end{equation}
\noindent
where \(\phi_{l}\) is the activation function. We will define the \emph{pseudo-gradient} operator as:
\begin{align}
\mathcal{D}v \coloneqq \nabla_{v}\,\log p_{\params}(\hat{\mathbf{y}} \mid \mathbf{x})
\end{align}
\noindent
for notational convenience. Notice that $\mathcal{D}v$ is a random variable whose randomness arises from sampling $\hat{\mathbf{y}}$. Using the properties of the Kronecker product, we can write the pseudo-gradient of $\params_{l}$ as:
\begin{align}
\mathcal{D}\params_{l} = \mathrm{vec}(\mathcal{D}\overline{\mathbf{W}}_{l})
= \mathrm{vec}(\mathcal{D}\mathbf{s}_{l} \overline{\mathbf{a}}_{l-1}^{\top}) = \overline{\mathbf{a}}_{l-1} \otimes  \mathcal{D}\mathbf{s}_{l},
\end{align}
where $\otimes$ is the Kronecker product. Then the $l$th block of the FIM can be computed as:
\begin{align}
\mathbf{F}_{l} &= \mathbb{E} \bigl[\mathcal{D} \params_{l} \mathcal{D}\params_{l}^{\top} \bigr] \\
&=\mathbb{E}\bigl[(\overline{\mathbf{a}}_{l-1}\otimes \mathcal{D} \mathbf{s}_{l})(\overline{\mathbf{a}}_{l-1}\otimes \mathcal{D} \mathbf{s}_{l})^\top  \bigr] \\
&= \mathbb{E}\bigl[\overline{\mathbf{a}}_{l-1}\,\overline{\mathbf{a}}_{l-1}^{\top} \otimes \mathcal{D} \mathbf{s}_{l} \mathcal{D} \mathbf{s}_{l}^{\top} \bigr] \\
&\approx
\mathbb{E}
\bigl[\overline{\mathbf{a}}_{l-1}\,\overline{\mathbf{a}}_{l-1}^{\top}\bigr] \otimes \mathbb{E} \bigl[\mathcal{D} \mathbf{s}_{l} \mathcal{D} \mathbf{s}_{l}^{\top}\bigr] \label{eqn:expectation_factorize} \\
&=\mathbf{A}_{l-1}\;\otimes\;\mathbf{S}_{l}, 
\end{align}
where have applied Kronecker product identities on the third equality.
Our final approximation $\widehat{\mathbf{F}}$ to the FIM is the block-diagonal matrix in which each block is $\mathbf{F}_{l}$. Here, $\mathbf{A}_{l-1} = \mathbb{E}[\overline{\mathbf{a}}_{l-1}\,\overline{\mathbf{a}}_{l-1}^{\top}] $ and $\mathbf{S}_{l} = \mathbb{E}[\mathcal{D}\mathbf{s}_{l} \mathcal{D}\mathbf{s}_{l}^{\top}]$ are the uncentered covariance matrices of the activations and the pre-activation pseudo-gradients with dimensions \((I+1)\times(I+1)\) and \(O\times O\), respectively. Practically, we can estimate the expectations via an empirical estimate and store the resulting statistics $\widehat{\mathbf{A}}_{l-1} = \frac{1}{N} \sum\limits_{\mathcal{D}} \overline{\mathbf{a}}_{l-1}\,\overline{\mathbf{a}}_{l-1}^{\top}$ and $\widehat{\mathbf{S}}_{l} = \frac{1}{N}\sum\limits_{\mathcal{D}}\mathcal{D}\mathbf{s}_{l} \mathcal{D}\mathbf{s}_{l}^{\top}$, and we define $\widehat{\mathbf{F}}_l \coloneq \widehat{\mathbf{A}}_{l-1} \otimes \widehat{\mathbf{S}}_l$.

To approximate $(\mathbf{F} + \lambda \mathbf{I})^{-1}\mathbf{v}$ for a vector $\mathbf{v}$ as needed to compute influence functions, we can compute $(\widehat{\mathbf{F}}_{l} + \lambda \mathbf{I})^{-1} \mathbf{v}_{l}$ separately for each layer $l$. Let \(\overline{\mathbf{V}}_{l}\) be the slice of \(\mathbf{v}\) reshaped to match \(\overline{\mathbf{W}}_{l}\), and define \(\mathbf{v}_{l} = \mathrm{vec}(\overline{\mathbf{V}}_{l})\). Applying the eigendecompositions $\mathbf{A}_{l - 1} = \mathbf{Q}_{\mathbf{A}_{l - 1}} \mathbf{D}_{\mathbf{A}_{l - 1}} \mathbf{Q}_{\mathbf{A}_{l - 1}}^\top $ and $\mathbf{S}_{l} = \mathbf{Q}_{\mathbf{S}_{l}} \mathbf{D}_{\mathbf{S}_{l}} \mathbf{Q}_{\mathbf{S}_{l}}^\top $, and using the Kronecker identity $\mathbf{U} \otimes \mathbf{V} = (\mathbf{Q}_\mathbf{U} \otimes \mathbf{Q}_\mathbf{V})\,(\mathbf{D}_\mathbf{U} \otimes \mathbf{D}_\mathbf{V})\,(\mathbf{Q}_\mathbf{U}^\top \otimes \mathbf{Q}_\mathbf{V}^\top)$ for two symmetric matrices $\mathbf{U} \otimes \mathbf{V}$, we can write:
\begin{align}
(\widehat{\mathbf{F}}_{l} + \lambda \mathbf{I})^{-1} \mathbf{v}_{l} &= (\mathbf{A}_{l-1} \otimes \mathbf{S}_{l} + \lambda \mathbf{I})^{-1} \mathbf{v}_{l} \\
&= (\mathbf{Q}_{\mathbf{A}_{l - 1}} \otimes \mathbf{Q}_{\mathbf{S}_{l}}){(\underbrace{\mathbf{D}_{\mathbf{A}_{l - 1}} \otimes \mathbf{D}_{\mathbf{S}_{l}}}_{\text{Scaling matrix $\mathbf{\Lambda}_{\text{KFAC}}$}} + \lambda \mathbf{I}_{\mathbf{A}_{l - 1}} \otimes \mathbf{I}_{\mathbf{S}_{l}})}^{-1} {\underbrace{(\mathbf{Q}_{\mathbf{A}_{l - 1}} \otimes \mathbf{Q}_{\mathbf{S}_{l}})}_{\text{Orthonormal eigenbasis}}}^\top \mathbf{v}_{l} \label{eqn:kfac}
\end{align}
where $\mathbf{I}_{\mathbf{A}_{l - 1}} $ and $\mathbf{I}_{\mathbf{S}_{l}}$ represent identity matrices of the same shape as $\mathbf{A}_{l - 1}$ and $\mathbf{S}_{l}$ respectively. 

\paragraph{Eigenvalue Corrected Kronecker-Factored Approximate Curvature}
One simple adjustment to the KFAC approximation can yield material improvements to the curvature approximation as well as the influence approximation. The KFAC formulation in \Cref{eqn:kfac} suggests that after expressing $\mathbf{v}_{l}$ in the eigenbasis $(\mathbf{Q}_{\mathbf{A}_{l - 1}} \otimes \mathbf{Q}_{\mathbf{S}_{l}})^\top$ we scale each element with $(\mathbf{D}_{\mathbf{A}_{l - 1}} \otimes \mathbf{D}_{\mathbf{S}_{l}} + \lambda \mathbf{I}_{\mathbf{A}_{l - 1}} \otimes \mathbf{I}_{\mathbf{S}_{l}})^{-1}$. Observing that for any matrix $\mathbf{W} = \mathbb{E}[\mathbf{u}\mathbf{u}^\top] = \mathbf{U}\mathbf{S}\mathbf{U}^\top$, it is true that $\mathbf{S}_{ii} = \mathbb{E}[(\mathbf{U}^\top \mathbf{v})_i^2]$, \citet{george2018fast} propose that a better scaling matrix is the diagonal matrix $\mathbf{\Lambda}_{\text{EKFAC}}$ with entries:
\begin{align}
\mathbf{\Lambda}_{ii} = \mathbb{E}\!\bigl[\bigl((\mathbf{Q}_{\mathbf{A}_{l -1}} \otimes \mathbf{Q}_{\mathbf{S}_{l}})^\top \mathcal{D}\params_l \bigr)^{2}_{i}\bigr].
\end{align}
In practice, this eigenvalue correction results in better influence estimates compared with KFAC.

\paragraph{Assumptions in \ekfac} From the equations above, we can see that KFAC \space makes two critical approximations which \ekfac \space inherits: First, it assumes that correlations between $\mathcal{D} \params_i$ and $\mathcal{D} \params_j$ are zero if they belong to different layers, yielding a block-diagonal approximation of $\mathbf{F}$. Second, it treats the activations as independent from the pre-activation pseudo-gradients, the basis of \Cref{eqn:expectation_factorize}. Furthermore, the matrix $\mathbf{S}_{l}$ depends on the \textit{sampled} labels $\widehat{\mathbf{y}}$, which for efficiency reasons is usually only sampled once per input $\inp$ which also may introduce some approximation error. In total, these assumptions cause $\mathbf{F}_{\text{EKFAC}}$ to differ from the true FIM $\mathbf{F}$, which is an error that \algoname \space corrects. \citet{grosse2016kronecker} introduce the KFAC approximation for convolution layers which introduces two further assumptions -- spatially uncorrelated derivatives and spatial homogeneity. Consistent with past works \citep{bae2024training, choe2024your}, we find that \ekfacif \space struggles for convolution architectures in comparison with MLPs. \algonameif's TDA performance on \resnetnine \space achieves a particularly large improvement compared to \ekfacif.

\paragraph{Cost of Computing \ekfacif} The three main components of computing \ekfacif \space are: 1) collecting the statistics $\widehat{\mathbf{A}}_{l-1}$ and $\widehat{\mathbf{S}}_{l}$, which requires a backward pass over the whole dataset. 2) computing the eigendecompositions $\mathbf{A}_{l - 1} = \mathbf{Q}_{\mathbf{A}_{l - 1}} \mathbf{D}_{\mathbf{A}_{l - 1}} \mathbf{Q}_{\mathbf{A}_{l - 1}}^\top $ and $\mathbf{S}_{l} = \mathbf{Q}_{\mathbf{S}_{l}} \mathbf{D}_{\mathbf{S}_{l}} \mathbf{Q}_{\mathbf{S}_{l}}^\top$, whose total precise cost depends on the architecture \citep{grosse2023studyinglargelanguagemodel}. 3) if we want to search the whole dataset $\mathcal{D}$ for influential examples, once we approximate $(\mathbf{F} + \lambda \mathbf{I})^{-1}\querygradwithfinalparamsnorand$ via the procedure above, we need to take a dot product with every training example gradient under consideration in $\mathcal{D}$. \algoname \space adds an incremental iterative procedure to the cost of \ekfacif \space which results in stronger TDA performance.

\section{Introducing \algonamesource}\label{appendix:astra_source}
\paragraph{Understanding the \ihvp} We have discussed how to compute \ihvps \space and therefore influence functions with \algoname. We can also apply \algoname \space to \source \space by making the substitution $\overline{\mathbf{r}}_{\ell} \approx \left( \overline{\ggn}_{\ell} + 
 \overline{\eta}_{\ell}^{-1} K_{\ell}^{-1} \eye \right)^{-1} \overline{\mathbf{g}}_{\ell} \coloneq \tilde{\mathbf{r}}_{\ell}$ described in \Cref{appendix:unrolled}, which replaces the matrix exponential in $\overline{\mathbf{r}}_{\ell}$. To understand the approximation, observe that the following term (with some rearranging) found in \Cref{eqn:source_approximations} $\overline{\LR}_{\ell} \sum_{k=T_{\ell-1}}^{T_{\ell}-1}  (\eye - \overline{\eta}_{\ell} \overline{\ggn}_{{\ell}})^{T_{\ell}-1-k} \overline{\mathbf{g}}_{\ell}$ resembles applying the Neumann series iterations on the vector $\mathbf{v} = \overline{\mathbf{g}}_{\ell}$ and the matrix $\mathbf{A} = \overline{\ggn}_{\ell}$ with a learning rate of $\overline{\LR}_{\ell}$ for a total of $K_{\ell}-1$ iterations. For large enough $K_{\ell}$, the truncation can be seen as approximately running Neumann series iterations until convergence, which results in the \ihvp \space $\overline{\ggn}_{\ell} ^{-1} \overline{\mathbf{g}}_{\ell}$. However, each segment actually only involves $K_{\ell}$ iterations at an average learning rate of $\overline{\LR}_{\ell}$, and thus is more closely related to \textit{truncated} Neumann series iterations (\Cref{appendix:truncated_neumann_series}), in which the parameters make less progress in the low eigenvalue directions. \citet{bae2024training} provide an interpretation that this can be approximated with damping as follows: $\overline{\LR}_{\ell} \sum_{k=T_{\ell-1}}^{T_{\ell}-1}  (\eye - \overline{\eta}_{\ell} \overline{\ggn}_{{\ell}})^{T_{\ell}-1-k} \overline{\mathbf{g}}_{\ell} \approx (\overline{\ggn} + {\overline{\eta}_{\ell}^{-1} K_{\ell}^{-1}} \eye)^{-1} \overline{\mathbf{g}}_{\ell}$, and that over a wide range of eigenvalues, the qualitative behavior matches well. The transformation of this term into an \ihvp \space allows us to apply \algoname \space to \source.
 
\paragraph{Practical Implementation} Recall that the final goal in \source \space is to approximate $\querygradwithfinalparamsnorand^\top\E _{\xi_b} \left[ \frac{\mathrm{d} \params_{T}}{\mathrm{d} \epsilon} \right]$. To do this, we follow \Cref{eqn:source_approximations_final} from left to right: for all $\ell = 1, \ldots, L$ segments, we first compute  $-\frac{1}{N} \querygradwithfinalparamsnorand^\top \prod_{\ell'=L}^{\ell+1} \overline{\segment}_{\ell'}$ in the same manner as \source. This will be the vector in our \ihvp. \algoname \space differs from \source \space in that instead of multiplying this vector by another matrix exponential, we multiply it by the inverse damped GGN $\left( \overline{\ggn}_{\ell}
 + \overline{\eta}_{\ell}^{-1} K_{\ell}^{-1} \eye \right)^{-1}$, which we can do using \algoname \space in the same manner described previously. Finally, we multiply by the average gradient, $\overline{\mathbf{g}}_{\ell}$ and accumulate the result over $L$ segments, which is done in the same manner as \source. Compared to \algonameif, there is an additional detail introduced: how to obtain the average GGN $\overline{\ggn}_{\ell}$. There are a number of options available, but we find that using the average weights in the segment works well, which we discuss below. The full procedure of applying \algoname \space to \source \space requires $L$ \ihvps \space per query. In many cases, the number of segments $L$ is likely to be small.\footnote{Part of the motivation for \source \space is to devise a scalable TDA algorithm for multi-stage training procedures, in which the number of segments is likely to be modest.  \citet{bae2024training} use $L=3$. } Since the preconditioners used by \algoname \space would need to be computed if we were to run \ekfacsource \space anyways, the incremental cost of \algonamesource \space compared to \ekfacsource \space is no more than $L$ times the number of iterations for each \ihvp, which is hundreds of iterations in our experiments. 

\paragraph{Approximation of Other Terms} We have discussed how to improve the approximation of $\overline{\mathbf{r}}_{\ell}$ with \algoname, but have not discussed $\overline{\segment}_{\ell'}$. We found evidence that improving the quality of the term $\overline{\mathbf{r}}_{\ell}$ improved TDA performance, but could not find the same evidence for $\overline{\segment}_{\ell'}$. Therefore we spent additional compute on improving the \ihvp \space approximation. As a result, we will leave $\overline{\segment}_{\ell'}$ as the approximation involving \ekfac.

\paragraph{Computing the Average Gauss-Newton Hessian} There are a number of options in computing the average GGN $\overline{\ggn}_{\ell}$.
\textbf{Option 1)}: since $\overline{\ggn}_{\ell} \coloneqq \mathbb{E}_{\xi_b}[\mathbf{\ggn}_k]$ for $T_{\ell-1} \leq k < T_{\ell}$, one can compute the matrix-vector product involving $\overline{\ggn}_{\ell}$ and $\mathbf{v}\in\mathbb{R}^D$ simply by taking an empirical average over samples within the segment: $\overline{\ggn}_{\ell}\mathbf{v} \approx \frac{1}{K_\ell}\sum_{T_{\ell-1} \leq k < T_{\ell}} \mathbb{E}_{\xi_b}[\ggn_k] \mathbf{v}$. This might be costly since one would have to load multiple checkpoints into memory just to compute a forward pass. \textbf{Option 2)}: Instead of sampling \textit{every} checkpoint in the segment, we could reduce the number of samples and take the empirical average of that instead. This is consistent with SOURCE as it uses only a subset of checkpoints in each segment anyways. If the segments chosen in SOURCE are indeed stationary as the derivation approximates, then in the extreme, we could take one checkpoint in each segment as the representative.
\textbf{Option 3)}: \citet{bae2024training} provides an alternative averaging scheme called FAST-SOURCE, in which one averages the parameters rather than the gradients, and shows that it works approximately as well as SOURCE. We tested Option 2 and Option 3 for a few settings and could not find meaningful differences, so opted to present the results for Option 3, using the average weights.

\section{Extended Related Works}\label{appendix:extended_related}

\paragraph{Understanding Influence Functions} A number of previous works \citep{basu2020influence, bae2022if, sogaard2021revisiting, saunshi2022understanding, epifano2023revisiting,  hu2024most, schioppa2024theoretical} have studied influence functions accuracy in various modern neural network settings. \cite{basu2020influence} show that influence functions do not approximate LOO retraining well. \cite{bae2022if} discover that the derivation of influence functions actually approximate the \textit{Proximal Bregman Response Function} (PBRF) rather than LOO retraining, which can be seen as an objective which tries to \textit{maximize} loss on the removed training example, subject to constraints in function space and weight space measured from the final parameters. Two large contributions to influence functions error are the \textit{warm-start retraining} assumption, which assumes that the counterfactual model is initialized at the final parameters, and the \textit{non-convergence gap}, which relates to the fact that the derivation assumes the model has converged to an optimal solution. Ensembling can help address the warm-start retraining assumption, while the non-convergence gap is addressed by \cite{bae2024training}. Others have focused on whether influence functions can accurately approximate group influence \citep{koh2019accuracy, basu2020second, saunshi2022understanding, ilyas2022datamodels} as it makes a linearity assumption due to the fact that it is a first-order approximation. While LOO influence is very noisy \citep{nguyen2024bayesian, bae2024training}, \citet{ilyas2022datamodels} discover that model predictions are approximately linear with respect to training example inclusion, which provides the justification for LDS \citep{park2023trak}. Overall, these works typically aim to address the fundamental assumptions surrounding influence functions (\Cref{tab:influence_assumptions}). In contrast, our work shows that a poorly approximated \ihvp \space can cause substantial performance degradation.

\paragraph{Ensembling in Influence Functions} Ensembling combines multiple models for improved generalization, uncertainty estimation, and calibration. It is a common approach to estimate the model posterior $p(\param | \mathcal{D})$ in Bayesian deep learning. Different strategies for sampling models as members of an ensemble exist: For example, deep ensembles sample models from varying random initializations~\citep{lakshminarayanan2017simple} to represent variations stemming from possible training trajectories, while stochastic weight averaging (SWA) approaches sample model parameters from the final iterations of model training~\citep{izmailov2018averaging, maddox2019simple}, which has the advantage of reduced training cost. Other methods may combine these ideas~\citep{wilson2020bayesian}, or use different approaches (e.g., Dropout~\citep{gal2016dropout}). 
Taking the average across several runs with varying sources of training process randomness is a common approach to account for the variability of model training in TDA estimation \citep{ilyas2022datamodels, park2023trak, bae2024training}. This can be seen as sampling from the distribution of true TDA to estimate the average treatment effect of excluding a training subset~\citep{nguyen2024bayesian} and has been effective in stabilizing estimations as well as evaluations (e.g. the LDS~\citep{park2023trak}). The size of the ensemble is generally connected to improved estimation quality of TDA scores, as shown in \cite{park2023trak}. Our results show that \algoname \space enjoys a larger performance boost from ensembling.

\section{Evaluating TDA}\label{appendix:evaluating_tda}

In this paper, we evaluate the performance of TDA methods with the LDS~\citep{park2023trak}, a widely used metric which we shortly described in \Cref{section:experiments}. Besides LDS, other evaluations for measuring TDA performance also exist. In this section, we present alternate methods of TDA evaluation. 

\paragraph{Expected leave-one-out retraining.} 
TDA methods usually define the influence of a training sample $\dataPoint_m$ on the model as the change in the model's predictions if $\dataPoint_m$ were not part of the training set. Hence, a straightforward way to compute the ground-truth to compare against is leave-one-out (LOO) retraining, as done in \cite{koh2017understanding, guu2023simfluence, yeh2018representer}. However, since the stochasticity inherent to model training makes LOO a noisy measure~\citep{epifano2023revisiting, nguyen2024bayesian}, the LOO score should be considered in expectation over the training process stochasticity $\xi$. We can then define the expected leave-one-out (ELOO) score as:
 \begin{align}
    \text{ELOO} (\dataPoint_m, \dataPoint_q, \mathcal{D}) \coloneqq \mathbb{E}_\xi [f_{\dataPoint_q}(\finalParams(\mathcal{D} \setminus \{\dataPoint_m\}; \xi))] - \mathbb{E}_\xi [f_{\dataPoint_q}(\finalParams (\mathcal{D};  \xi))].
\end{align}
This score can be viewed as the ground-truth average treatment effect (ATE) \citep{holland1986statistics} of the removal of $\dataPoint_m$ from $\mathcal{D}$. While principled, in practice, the effect of removing a single point is highly noisy~\citep{bae2024training, nguyen2024bayesian} so that a stable estimate of ELOO may only be achieved with an extremely large number of samples to compute the empirical expectation. 
In contrast, LDS considers the ATE of excluding a \emph{group} of training samples from training, which has been shown to be more stable in expectation~\citep{park2023trak, bae2024training}.

\paragraph{Top-k removal and retraining.}
The sign of the TDA scores indicates whether the excluded training samples are positively or negatively influential~\citep{bae2024training}, also referred to as proponents and opponents~\citep{pruthi2020estimating}, helpful and harmful samples~\citep{koh2017understanding}, or excitatory and inhibitory~\citep{yeh2018representer}, respectively \cite{bae2024training}. The idea behind this evaluation is that the removal of positively influential samples $\{\dataPoint_m\}$ removes support for the query sample $\dataPoint_q$ and consequently should lead to a change in prediction confidence on $\dataPoint_q$~\citep{han2020explaining}. 
\cite{han2020explaining} conduct this evaluation by removing the top and bottom 10\% of training samples ranked by influence functions and compare against the removal of the least influential (i.e., smallest influence scores by magnitude) and random samples to see if the resulting models change their predictions in the expected way.
Similar to top-k removal and retraining, previous work has tested how many highly influential samples need to be removed to \textit{flip} a prediction~\citep{yang2023many, bae2024training}, called subset removal counterfactual evaluation. Similar to LDS, removing top-k samples is based on counterfactual retraining with excluded groups. However, one core difference is that since LDS involves a sum over all the attribution scores for every $\dataPoint_m$ in each subset (\Cref{eqn:lds_definition}), poorly \textit{calibrated} attribution scores resulting in one outlier score $\tau(\dataPoint_m, \dataPoint_q, \mathcal{D})$ may result in poor LDS, demanding that the TDA method assigns calibrated scores across all points $\dataPoint_m$ in consideration.

\section{Experiment Details}\label{appendix:experiment_details}

\subsection{Choice of Measurement Function}\label{subsection:measurement}
While in principle the derivation of our influence scores does not restrict what measurement function $f_{\dataPoint_q}$ is used, in practice some choices of measurement functions work better than others. In our experiments, for regression problems, we use the measurement function:
\begin{align}
f_{\dataPoint_q}(\params) = \lvert g(\params, \mathbf{x}_q) - \mathbf{t}_q \rvert
\end{align}
where $g(\params, \mathbf{x}_q)$ denotes the last layer output of the neural network when $\mathbf{x}_q$ is the input and $\mathbf{t}_q$ is a scalar output. For all classification problems with the exception of \gpttwo, we use the measurement function:
\begin{align}
f_{\dataPoint_q}(\params) &= - \log \frac{\sigma(g(\params, \mathbf{x}_q))_{\mathbf{t}_q}}{1-\sigma(g(\params, \mathbf{x}_q))_{\mathbf{t}_q}} \\ &= - g(\params, \mathbf{x}_q)_{\mathbf{t}_q} + \log \left( \sum_{i} \exp g(\params, \mathbf{x}_q)_i - \exp g(\params, \mathbf{x}_q)_{\mathbf{t}_q} \right) ,
\end{align}
where $\sigma$ denotes the softmax function and the subscript $\mathbf{t}_q$ refers to the taking the entry corresponding to the correct class. This measurement function is identical to the one found in \cite{park2023trak}. For \gpttwo, we use the training loss as the measurement function.

\subsection{Experiment Details}\label{subsection:hyperparameters}

\Cref{table:setting_details} shows the architecture and hyperparameters used to compute the final parameters $\params^s$, which we use to run \ekfacif, \ekfacsource, \algonameif \space and \algonamesource. The first column shows the size of the training dataset and the number of query examples in each setting. The third column shows the hyperparameters used to train the models in each setting; we use the same hyperparameters as reported in \citet{bae2024training}. We estimate the expected ground-truth in the left-hand-side of \Cref{eqn:lds_definition} with an empirical average -- the last column in \Cref{table:setting_details} shows the number of repeats \textit{per mask} $\mathcal{S}_j$ (i.e., number of $\xi$ sampled) to estimate this value. All settings use a constant learning rate, with the exception of \cifarten, which uses a cyclic learning rate schedule.

\begin{table}[h]
    \centering
    \captionsetup{font=small}
    \caption{\textbf{\small{Summary of training details.}}}
    \label{table:setting_details}
    \resizebox{\columnwidth}{!}{
    \begin{tabular}{@{}ccccc@{}}
    \toprule
    \textbf{Dataset} & \textbf{Architecture} & \textbf{Hyperparameters} & \textbf{Ground-truth Retraining} \\ \midrule
    \begin{tabular}[c]{@{}c@{}} \textbf{UCI Concrete} \\ 896 training examples \\ 103 query examples \end{tabular}
    & \begin{tabular}[c]{@{}c@{}} MLP - 4 Layers \\ (128, 128, 128) Hidden Units\\ ReLU activation \end{tabular}  & \begin{tabular}[c]{@{}c@{}} SGD w/ momentum \\ Learning rate: $3 \times 10^{-2}$\\ Weight decay: $10^{-5}$  \\Momentum: $0.9$ \\ Batch size: $32$ \\ Epochs: $20$ \end{tabular} &  \begin{tabular}[c]{@{}c@{}} Repeats: 100 \\ Masks: 100 \\ Total: 10,000 \end{tabular} \\ \midrule
    \begin{tabular}[c]{@{}c@{}} \textbf{UCI Parkinsons} \\ 5,280 training examples \\ 100 query examples \end{tabular}        & \begin{tabular}[c]{@{}c@{}} MLP - 4 Layers \\ (128, 128, 128) Hidden Units\\ ReLU activation \end{tabular} & \begin{tabular}[c]{@{}c@{}} SGD w/ momentum \\ Learning rate: $10^{-2}$ \\ Weight decay: $3 \times 10^{-5}$  \\ Momentum: $0.9$ \\ Batch size: $32$ \\ Epochs: 20 \end{tabular} &  \begin{tabular}[c]{@{}c@{}} Repeats: 100 \\ Masks: 100 \\ Total: 10,000 \end{tabular} \\ \midrule
    \begin{tabular}[c]{@{}c@{}} \textbf{MNIST (Subset)} \\ 6,144 training examples \\ 100 query examples \end{tabular}         &  \begin{tabular}[c]{@{}c@{}} MLP - 4 Layers \\ (512, 256, 128) Hidden Units\\ ReLU activation \end{tabular} & \begin{tabular}[c]{@{}c@{}} SGD w/ momentum \\ Learning rate: $3 \times 10^{-2}$  \\ Weight decay: $10^{-3}$ \\ Momentum: $0.9$ \\ Batch size: $64$ \\Epochs: $20$  \end{tabular} &  \begin{tabular}[c]{@{}c@{}} Repeats: 50 \\ Masks: 100 \\ Total: 5,000 \end{tabular}
       \\ \midrule
    \begin{tabular}[c]{@{}c@{}} \textbf{\fashionmnist \space (Subset)} \\ 6,144 training examples \\ 100 query examples \end{tabular}  & \begin{tabular}[c]{@{}c@{}} MLP - 4 Layers \\ (512, 256, 128) Hidden Units\\ ReLU activation \end{tabular}  & \begin{tabular}[c]{@{}c@{}} SGD w/ momentum \\ Learning rate: $3 \times 10^{-2}$ \\ Weight decay: $10^{-3}$  \\Momentum: $0.9$ \\ Batch size: $64$ \\ Epochs: $20$  \end{tabular}  &  \begin{tabular}[c]{@{}c@{}} Repeats: 50 \\ Masks: 100 \\ Total: 5,000 \end{tabular}
       \\  
    \midrule
    \begin{tabular}[c]{@{}c@{}} \textbf{CIFAR-10 (Subset)} \\ 3,072 training examples \\ 100 query examples \end{tabular}           & ResNet-9  \citep{he2016deep} & \begin{tabular}[c]{@{}c@{}} SGD w/ momentum \\ Peak learning rate: $0.4$ \\    Weight decay: $10^{-3}$ \\Momentum: $0.9$ \\ Batch size: $512$ \\Epochs: 25 \end{tabular} &  \begin{tabular}[c]{@{}c@{}} Repeats: 50 \\ Masks: 100 \\ Total: 5,000 \end{tabular}
       \\
    \midrule
    \begin{tabular}[c]{@{}c@{}} \textbf{\wikitexttwo} \\ 4,656 training sequences \\ 512 sequence length \\ 100 query sequences \end{tabular}           & \gpttwo \space  \citep{radford2019language}  & \begin{tabular}[c]{@{}c@{}} AdamW \\ Learning rate: $3 \times 10^{-5}$\\ Weight decay: $10^{-2}$  \\ Batch size: $8$ \\ Epochs: 3 \end{tabular}  &  \begin{tabular}[c]{@{}c@{}} Repeats: 10 \\ Masks: 100 \\ Total: 1,000
    \end{tabular}\\
    \midrule
    \begin{tabular}[c]{@{}c@{}} \textbf{\fashionmnist-N} \\ 6,144 training examples \\ 100 query examples \\ 30\% of the dataset randomly relabeled \end{tabular}           & \begin{tabular}[c]{@{}c@{}} MLP - 4 Layers \\ (512, 256, 128) Hidden Units\\ ReLU activation \end{tabular}  & \begin{tabular}[c]{@{}c@{}} SGD w/ momentum \\ Learning rate: $10^{-2}$\\ Weight decay: $3 \times 10^{-5}$ \\ Momentum: $0.9$  \\ Batch size: $64$ \\ Epochs: 3 \end{tabular}  &  \begin{tabular}[c]{@{}c@{}} Repeats: 50 \\ Masks: 100 \\ Total: 5,000
    \end{tabular} \\
    \bottomrule
    \end{tabular} 
    }
\end{table}

\Cref{table:experiment_details} shows the details for the various TDA algorithms we use in our experiments.

For \ekfacif, we use the same damping as the corresponding \algonameif \space for comparability. \source \space provides a natural value for damping from its derivation: $\lambda_{\ell} = 1/\overline{\eta}_{\ell} K_{\ell}$ \citep{bae2024training}, allowing us to sidestep tuning the damping term. For both \ekfacif \space and \algonameif, we use the damping value implied by \source \space for comparability between influence functions and \source: we take the  average $\lambda_{\ell}$ implied by \source \space for each segment and weigh them by the total iterations $K_{\ell}$ in segment $\ell$ as our influence functions damping value.

\ekfacif, \ekfacsource, \algonameif, and \algonamesource \space all compute influence on the same set of layers. For MLP architectures, we compute influence on all layers. For \resnetnine, we compute influence on MLP and convolution layers. For \gpttwo, we compute influence only on MLP layers in line with \citet{grosse2023studyinglargelanguagemodel}.

\begin{center}
\small  %
\begin{longtable}[h]{@{}
  >{\centering\arraybackslash}m{.18\columnwidth}  %
  >{\raggedright\arraybackslash}m{.4\columnwidth} %
  >{\raggedright\arraybackslash}m{.4\columnwidth} %
  @{}}
  \caption{\textbf{Summary of TDA Algorithm Details}}
  \label{table:experiment_details}
  \\ \toprule
    \textbf{Dataset}
  & \textbf{\algonameif\  Details}
  & \textbf{\algonamesource\  Details}
  \\ \midrule
  \endfirsthead

  \multicolumn{3}{@{}l}{\small\sl continued from previous page}\\
  \toprule
    \textbf{Dataset}
  & \textbf{\algoname\ Influence Function Details}
  & \textbf{\algoname\ SOURCE Details}
  \\ \midrule
  \endhead

  \midrule
  \multicolumn{3}{r}{\small\sl continued on next page}\\
  \endfoot

  \bottomrule
  \endlastfoot

  \textbf{UCI Concrete}
  & We run \algoname \space for 200 iterations, and use GGN damping factor $\lambda=0.0017$, preconditioner damping factor $\tilde{\lambda}=0.0017$, learning rate $\alpha=0.1\lambda$, batch‑size 256, SGD w/ 0.9 momentum, and decay the learning rate by 0.5 every 50 iterations.
  & We use 3 segments. For each segment, we run \algoname \space for 200 iterations, and use preconditioner damping factor equal to the GGN damping factor $\tilde{\lambda}_{\ell} = \lambda_{\ell} = 1/\overline{\eta}_{\ell} K_{\ell}$, learning rate $\alpha_{\ell} = 0.1\lambda_{\ell}$, batch‑size 256, SGD w/ 0.9 momentum, and decay the learning rate by 0.5 every 50 iterations.
  \\
  \\ \midrule

  \textbf{UCI Parkinsons}
  & We run \algoname \space for 200 iterations, and use GGN damping factor $\lambda=0.00091$, preconditioner damping factor $\tilde{\lambda}=0.00091$, learning rate $\alpha=0.1\lambda$, batch‑size 256, SGD w/ 0.9 momentum, and decay the learning rate by 0.5 every 50 iterations.
  & We use 3 segments. For each segment, we run \algoname \space for 200 iterations, and use preconditioner damping factor equal to the GGN damping factor $\tilde{\lambda}_{\ell} = \lambda_{\ell} = 1/\overline{\eta}_{\ell} K_{\ell}$, learning rate $\alpha_{\ell} = 0.1\lambda_{\ell}$, batch‑size 256, SGD w/ 0.9 momentum, and decay the learning rate by 0.5 every 50 iterations.
  \\ \midrule

  \textbf{\mnist}
  & We run \algoname \space for 200 iterations, and use GGN damping factor $\lambda=0.0052$, preconditioner damping factor $\tilde{\lambda}=0.0052$, learning rate $\alpha=0.1\lambda$, batch‑size 256, SGD w/ 0.9 momentum, and decay the learning rate by 0.5 every 50 iterations.
  & We use 3 segments. For each segment, we run \algoname \space for 200 iterations, and use preconditioner damping factor equal to the GGN damping factor $\tilde{\lambda}_{\ell} = \lambda_{\ell} = 1/\overline{\eta}_{\ell} K_{\ell}$, learning rate $\alpha_{\ell} = 0.1\lambda_{\ell}$, batch‑size 256, SGD w/ 0.9 momentum, and decay the learning rate by 0.5 every 50 iterations.
  \\ \midrule

  \textbf{\fashionmnist}
  & We run \algoname \space for 200 iterations, and use GGN damping factor $\lambda=0.0052$, preconditioner damping factor $\tilde{\lambda}=0.0052$, learning rate $\alpha=0.1 \lambda$, batch‑size 256, SGD w/ 0.9 momentum, and decay the learning rate by 0.5 every 50 iterations.
  & We use 3 segments. For each segment, we run \algoname \space for 200 iterations, and use preconditioner damping factor equal to the GGN damping factor $\tilde{\lambda}_{\ell} = \lambda_{\ell} = 1/\overline{\eta}_{\ell} K_{\ell}$, learning rate $\alpha_{\ell} = 0.1\lambda_{\ell}$, batch‑size 256, SGD w/ 0.9 momentum, and decay the learning rate by 0.5 every 50 iterations.
  \\ \midrule

  \textbf{\cifarten}
  & We run \algoname \space for 200 iterations, and use GGN damping factor $\lambda=0.014$, preconditioner damping factor $\tilde{\lambda}=0.014$, learning rate $\alpha=0.01 \lambda$, batch‑size 128, SGD w/ 0.9 momentum, and decay the learning rate by 0.5 every 50 iterations.
  & We use 3 segments. For each segment, we run \algoname \space for 200 iterations, and use preconditioner damping factor equal to the GGN damping factor $\tilde{\lambda}_{\ell} = \lambda_{\ell} = 1/\overline{\eta}_{\ell} K_{\ell}$, learning rate $\alpha_{\ell} = 0.01\lambda_{\ell}$, batch‑size 128, SGD w/ 0.9 momentum, and decay the learning rate by 0.5 every 50 iterations.
  \\ \midrule
  
  \textbf{\wikitexttwo}
  & We run \algoname \space for 300 iterations, and use GGN damping factor $\lambda=0.011$, preconditioner damping factor $\tilde{\lambda}=0.011$, learning rate $\alpha= \lambda$, batch‑size 16, SGD w/ 0.9 momentum, and decay the learning rate by 0.9 every 100 iterations.
  & We use 3 segments. For each segment, we run \algoname \space for 300 iterations, and use preconditioner damping factor equal to the GGN damping factor $\tilde{\lambda}_{\ell} = \lambda_{\ell} = 1/\overline{\eta}_{\ell} K_{\ell}$, learning rate $\alpha_{\ell} = \lambda_{\ell}$, batch‑size 8, SGD w/ 0.9 momentum, and decay the learning rate by 0.9 every 100 iterations.
  \\ \midrule

    \textbf{\fashionmnist-N}
  & We run \algoname \space for 200 iterations, and use GGN damping factor $\lambda=0.10$, preconditioner damping factor $\tilde{\lambda}=0.10$, learning rate $\alpha=0.1\lambda$, batch‑size 256, SGD w/ 0.9 momentum, and decay the learning rate by 0.5 every 50 iterations.
  & We use 3 segments. For each segment, we run \algoname \space for 200 iterations, and use preconditioner damping factor equal to the GGN damping factor $\tilde{\lambda}_{\ell} = \lambda_{\ell} = 1/\overline{\eta}_{\ell} K_{\ell}$, learning rate $\alpha_{\ell} = 0.1 \lambda_{\ell}$, batch‑size 256, SGD w/ 0.9 momentum, and decay the learning rate by 0.5 every 50 iterations.
  \\

\end{longtable}
\end{center}
\paragraph{Other Baselines} In \Cref{fig:ensembling_benefits}, in addition to the EKFAC-based baselines, we also compare \algoname \space against TracIn \citep{pruthi2020estimating} and TRAK \citep{park2023trak}. We use the same checkpoints for TracIn as \source \space for comparability. TRAK applies random projections to reduce the computation and memory footprint -- for \mnist, \fashionmnist(-N) and UCI Parkinsons, we use projection dimensions of 4,096 and for UCI Concrete, we use projection dimension of 512, consistent with \cite{bae2024training}. For \cifarten, we do a hyperparameter sweep among [128, 256, 512, 1024, 2048, 4096, 8192, 20480] for the best hyperparameter (512). We omit TRAK's results on \gpttwo \space due to lack of publicly available implementations.

\paragraph{Training Curve Details} In \Cref{fig:training_curves}, we compare \algonameif \space and \sni \space on an arbitrary $\dataPoint_q$, using 10 seeds and report mean values along with shaded regions representing 1 standard error. To compute both \algonameif \space and the baseline \sni \space in \Cref{fig:training_curves}, we conduct a hyperparameter sweep for the learning rate from $10^{0}$ to $10^{-5}$ in increments of one order of magnitude with momentum set to zero, and use the best learning rate based on the average training objective over the last ten iterations. Both \sni \space and \algonameif \space use the same damping values and batch sizes for \Cref{fig:training_curves}, as listed in \Cref{table:experiment_details} for comparability. The learning rates used were $10^{-2}$ for all settings for \algonameif, and the best learning rates for SNI were $1, 0.1, 0.01, 0.1$ for \mnist, \fashionmnist, \cifarten, and UCI Concrete respectively.

\paragraph{Eigendecomposition Details} In \Cref{fig:decomposition}, we compare the performance of influence functions computed with \algoname \space and \sni \space after projecting to various subspaces. This experiment uses a smaller damping of $10^{-4}$ for both MNIST and \fashionmnist \space to be able to discern the impact of directions of low curvature on influence functions performance. We use the same batch sizes as disclosed in \Cref{table:experiment_details} and no momentum. For \algonameif, we used a learning rate of $10^{-4}$ for both settings. For the baseline SNI, the learning rates were 0.1 and 0.01 respectively for \mnist \space and \fashionmnist, which we found to give the strongest LDS for the subspace corresponding to $S_5$.

\ifneurips
\paragraph{Compute Resources} A shared cluster was used (both internal and external), consisting of a mix of A6000 (48GB),  A100 (80GB) and H100 (80GB) GPUs, which were used to conduct all experiments. Computing the ground truth of the LDS experiments in \Cref{fig:fig1} is an expensive component of the overall compute to replicate the experiments. For the most expensive setting, fine-tuning \gpttwo \space with \wikitexttwo, computing ground-truth costs at most 500 hours of compute with the resources listed above. Similarly, running \ekfacif, \ekfacsource, \algonameif \space and \algonamesource \space for the \gpttwo \space setting is the most expensive of all the settings. Running \algonamesource \space for the \gpttwo \space setting costs at most 40 hours with the compute resources listed above.
\fi

\paragraph{Statistical Significance}
For LDS experiments, we provide error bars indicating 1 standard error for all 1 model predictions estimated with 10 seeds. For the training curve (\Cref{fig:training_curves}) and the eigendecomposition experiments (\Cref{fig:decomposition}), we show 1 standard error estimated with 10 seeds. The standard error is computed by taking the standard deviation $s$ of the computed metric, divided by $\sqrt{10}$.

\paragraph{Assets}
We use \texttt{pytorch} 2.5.0 and the publicly available \texttt{kronfluence} package for our experiments, which can be found at \url{https://github.com/pomonam/kronfluence}.

\section{Implications of Truncated Neumann Series}\label{appendix:truncated_neumann_series}
In \Cref{subsection:compute_ihvps}, we introduced the connection between \vni \space and the \ihvp \space approximation. Iterative \ihvp \space approximation algorithms like \lissa \space \citep{agarwal2017second} usually requires a large number of iterations to achieve good \ihvp \space approximations~\citep{koh2017understanding, grosse2023studyinglargelanguagemodel}. In practice, the number of iterations in the algorithm is usually set with the assumption that the approximation converges within the iterations (e.g., 5000 in \cite{koh2017understanding}). While a large number of iterations makes convergence likely, it is not a guarantee, and raising the number of iterations implies additional computational cost.
We demonstrate that using fewer iterations and stopping before convergence can be interpreted as adding an additional damping term to the iHVP approximation. There is existing work noting that the truncation of Neumann series can be viewed as increased damping \citep{vicol2022implicit}, but here we present it with the derivation technique found in \citep{bae2024training}.
We derive the following expression for the truncated Neumann series with $J$ iterations:
\begin{align}
    \alpha \sum_{j=0}^{J-1} \left( \mathbf{I} - \alpha \ggn - \alpha \lambda \mathbf{I} \right)^j &= (\ggn + \lambda \mathbf{I})^{-1} (\mathbf{I} - (\mathbf{I} - \alpha \ggn - \alpha \lambda \mathbf{I})^J) \label{eqn:geometric_series} \\
    &\approx (\ggn + \lambda \mathbf{I})^{-1} (\mathbf{I} - \exp(-\alpha J \ggn - \alpha J \lambda \mathbf{I})) \label{eqn:exponential} \\
    &\approx (\ggn + \hat{\lambda} \mathbf{I})^{-1} \label{eqn:qualitative_damping},
\end{align}
where $\hat{\lambda} = \lambda + \alpha^{-1} J^{-1}$. 
\Cref{eqn:geometric_series} and \Cref{eqn:exponential} utilize the finite series and the matrix exponential definition, respectively. Given that the matrices in \Cref{eqn:exponential} commute, we can express it as the following matrix function:
\begin{align}
    F (\sigma) \coloneqq \frac{1 - \exp(-\alpha J \sigma - \alpha J \lambda)}{\sigma + \lambda}.
\end{align}
Let $\mathbf{G} = \mathbf{Q} \mathbf{D} \mathbf{Q}^\top$ be the eigendecomposition, where $\sigma_j$ denotes the $j$th eigenvalue of $\ggn$.
The expression in \Cref{eqn:exponential} can be interpreted as applying the function $F(\sigma)$ to each eigenvalue $\sigma$ of $\ggn$. In high-curvature directions, this term asymptotically approaches $\sfrac{1}{\sigma + \lambda}$, whereas in low-curvature directions with small values of $\lambda$, it tends toward $\alpha J$.

The qualitative behavior of the function $F$ can be captured by $F_{\text{inv}}$, defined as:
\begin{align}
    F_{\text{inv}} (\sigma) \coloneqq \frac{1}{\sigma + \lambda + \alpha^{-1} J^{-1}} \label{eq:matrix_function}
\end{align}
Applying $F_{\text{inv}}$ to the Hessian results in approximating \Cref{eqn:qualitative_damping} with a modified damping term $\hat{\lambda} \coloneqq \lambda + \sfrac{1}{\alpha J}$. Hence, the truncated version can be interpreted as incorporating a larger damping term, by $\sfrac{1}{\alpha J}$. The implicitly larger damping term affects the iterative updates and adds additional difficulty to tuning the hyperparameters of iterative \ihvp \space solvers. In \algoname, we leverage the \ekfac \space approximation as a preconditioner for the \ihvp \space approximation to improve the conditioning of the problem, which mitigates this issue.

\section{Limitations \& Broader Impact}\label{appendix:limitations_and_broader_impact}
\paragraph{Limitations} We have addressed the problem of computing accurate \ihvps \space for TDA. As we outlined in  \Cref{influence_functions}, in addition to the computing the \ihvp, a large component of the cost in computing influence functions for both \ekfacif \space and \algonameif \space when scanning the dataset for highly influential examples is taking the dot product with all the training example gradients, which is an orthogonal but important issue that we do not address. We also noted in our experiments that in some cases, the bottleneck for influence function performance in TDA is not an inaccurate \ihvp, but other factors such as violating the fundamental assumptions involved in the influence functions derivation. In these cases, the benefits from an improved \ihvp \space approximation may not materialize until other bottlenecks are resolved. For example, the \fashionmnist \space experiments show a large increase in performance after ensembling relative to the 1-model scores obtained by an improved \ihvp \space approximation, which suggests that stochasticity in the training procedure may be the dominant factor hindering TDA performance. Finally, for \algonamesource, we present a way in which we can apply the \ihvp \space computation, but we leave to future work to explore various ways to compute the average GGN $\overline{\ggn}_{\ell}$, which may further improve performance.

\paragraph{Broader impact} Our work improves the accuracy of the \ihvp \space approximation for use in TDA. The algorithmic improvements we present do not have direct societal impact. However, improved TDA can provide insights into the relation between training data and model behavior. From this perspective, our work has similar broader impact to other work in TDA, sharing similar potential benefits -- namely, enhanced interpretability, transparency, and fairness in AI systems and share similar risks. Specifically, advancing TDA can help us understand models through the lens of training data, which can be applied to many domains such as building more equitable and transparent machine learning systems \citep{brunet2019understanding, wang2024fairif}, and investigating questions of intellectual property and copyright \citep{van2021memorization, mlodozeniec2024influence, mezzi2025owns}. On the flip side, TDA can also be used to maliciously to craft data poisoning attacks and could result in models with undesirable behavior. It is important that TDA algorithms are improved with mitigation of the risks.

\end{document}